\documentclass[subscriptcorrection,upint,varvw,barcolor=Goldenrod3,mathalfa={cal=euler},balance,hyphenate,french,nolists]{asmejour} %

\usepackage{bm}
\usepackage{amsmath}
\usepackage{caption}
\usepackage{graphicx}
\usepackage{float}
\usepackage{stfloats}
\usepackage{multirow}
\usepackage{makecell} 
\usepackage{subcaption}
\usepackage{lipsum}
\usepackage{algorithm}
\usepackage{algorithmicx}
\usepackage{graphicx}
\usepackage{xcolor}
\usepackage[noend]{algpseudocode}
\usepackage{hyperref}
\JourName{ASME Journal of Mechanical Design}%<=== change to the name of your journal

\begin{document}

% Change to your author name[s] and addresses, in the desired order of authors.
% First name, middle initial, last name
% Use title case (upper and lower case letters)
% Note usage below for corresponding author.

\SetAuthorBlock{Ying-Kuan Tsai}{Department of Mechanical Engineering,\\
   Northwestern University,\\
   Evanston, IL, 60208 \\
   email: yingkuan.tsai@northwestern.edu} 

\SetAuthorBlock{Vispi Karkaria}{Department of Mechanical Engineering,\\
   Northwestern University,\\
   Evanston, IL, 60208 \\
   email: vispikarkaria2026@u.northwestern.edu} 

\SetAuthorBlock{Yi-Ping Chen}{Department of Mechanical Engineering,\\
   Northwestern University,\\
   Evanston, IL, 60208 \\
   email: chenyp@u.northwestern.edu}

% To label one or more corresponding authors put "Name\CorrespondingAuthor". No space after "Name".
% An optional argument can be added if email is not in address block as
%      "Name\CorrespondingAuthor{write@to.me}"
% Can also include multiple emails and use the command more than once for multiple corresponding authors,
%      "Name\CorrespondingAuthor{write@to.him, write@to.her}"

\SetAuthorBlock{Wei Chen\CorrespondingAuthor}{Department of Mechanical Engineering,\\
   Northwestern University,\\
   Evanston, IL, 60208 \\
   email: weichen@northwestern.edu} 

%%% Change to your paper title. Can insert line breaks if you wish (otherwise breaks are selected automatically).
\title{Digital Twin-enabled Multi-generation Control Co-Design with Deep Reinforcement Learning}

%%% Change these to your keywords.  Keywords are automatically printed at the end of the abstract.
%%% This command must come BEFORE the end of the abstract.
%%% If you don't want keywords, omit the \keyword{..} command.

%% Abstract should be no more than 250 words
\begin{abstract}
Control Co-Design (CCD) integrates physical and control system design to improve the performance of dynamic and autonomous systems. Despite advances in uncertainty-aware CCD methods, real-world uncertainties remain highly unpredictable. Multi-generation design addresses this challenge by considering the full lifecycle of a product: data collected from each generation informs the design of subsequent generations, enabling progressive improvements in robustness and efficiency. Digital Twin (DT) technology further strengthens this paradigm by creating virtual representations that evolve over the lifecycle through real-time sensing, model updating, and adaptive re-optimization. This paper presents a DT-enabled CCD framework that integrates Deep Reinforcement Learning (DRL) to jointly optimize physical design and controller. DRL accelerates real-time decision-making by allowing controllers to continuously learn from data and adapt to uncertain environments. Extending this approach, the framework employs a multi-generation paradigm, where each cycle of deployment, operation, and redesign uses collected data to refine DT models, improve uncertainty quantification through quantile regression, and inform next-generation designs of both physical components and controllers. The framework is demonstrated on an active suspension system, where DT-enabled learning from road conditions and driving behaviors yields smoother and more stable control trajectories. Results show that the method significantly enhances dynamic performance, robustness, and efficiency. Contributions of this work include: (1) extending CCD into a lifecycle-oriented multi-generation framework, (2) leveraging DTs for continuous model updating and informed design, and (3) employing DRL to accelerate adaptive real-time decision-making.
\end{abstract}

\date{Version \versionno, \today}%% You can modify this information as desired. (you can find it in 'asmejour.cls')
							%% Putting \date{} will suppress any date.  
							%% If this command is omitted, date defaults to \today
							%% This command must come somewhere before \maketitle

\maketitle %% This command creates the author/title/abstract block. Essential!

%%%%%%%%%%%%%%%%%%%%%%%%%%%%%%%%%%%%%%%%%%%%%%%%%%%%%%%%%%%%%%%%%%%%%%%%%%%%%%%%%%%%%%%%%%%%%%%%%%%%%%%
%%%%%%%%%%%%%%%%%%%%%  End of fields to be completed. Now write! %%%%%%%%%%%%%%%%%%%%%%%%%%%%%%%%%%%%%%

\section{Introduction}

% Autonomous systems --> CCD --> active suspension (Allison's papers)
Autonomous systems are increasingly being adopted across various domains, from self-driving vehicles to robotics, where adaptability to uncertain environments is critical. Traditional design approaches often treat the physical system design and control policy separately, leading to suboptimal solutions. Control Co-Design (CCD) addresses this limitation by simultaneously optimizing both the physical system and its controller to improve performance~\cite{garcia2019control,allison2014special,tsai2022constraint}. For example, active suspension systems in vehicles offer significant advantages over passive suspension systems, which rely solely on mechanical properties to absorb road disturbances~\cite{allison2014co,bayat2023control,tsai2023phd}. Active suspension employs real-time control adjustments to enhance ride comfort, stability, and safety. By leveraging CCD, the design and control of active suspension can be jointly optimized, leading to superior performance in diverse and uncertain driving conditions.

% Challenges (environments and driving behaviors) --> uncertainty-aware CCD by assuming disturbances and model-based approaches --> problematic
However, achieving robust CCD in real-world applications presents several challenges. Vehicle suspension systems, for example, operate under highly unpredictable conditions, such as varying road surfaces, changing weather patterns, and diverse driver behaviors. Some existing methods focus on uncertainty-aware CCD to address these uncertainties, where they are explicitly considered during the design stage to generate robust solutions, including stochastic programming~\cite{bravo2022robust}, robust Model Predictive Control (MPC)~\cite{nash2021robust,tsai2023robust}, stochastic MPC~\cite{tsai2025control}, reliability-based design optimization~\cite{cui2021reliability}, and robust design optimization~\cite{rudnick2022robust,azad2023overview}. While these methods improve robustness, they typically assume well-defined disturbances and may struggle to adapt to unforeseen variations in real-world conditions. Additionally, these model-based approaches present a critical challenge in real-world applications, limiting their practical applicability. These limitations highlight the need for approaches that can go beyond predefined disturbance models and generalize across diverse operating conditions. Data-driven methods offer a promising alternative for solving CCD problems by leveraging real-world data to continuously update system models, capture uncertainties more effectively, and enhance adaptability in dynamic environments.

Numerous studies have integrated reinforcement learning (RL) techniques into CCD, enabling autonomous systems to co-optimize both their physical design and control policies based on performance feedback~\cite{chen2021co}. Unlike traditional model-based methods, RL-based CCD frameworks jointly optimize both physical structures and control policies by learning directly from interactions with the environment, making them well-suited for handling high-dimensional design spaces and adapting to changing conditions. Two major formulations have been explored. The first is bi-level optimization, where the high-level process searches for optimal hardware parameters while the low-level RL agent learns an optimal control policy through trial and error~\cite{sadat2023two}. To improve sample efficiency and exploration, techniques such as Bayesian optimization and evolutionary algorithms have been integrated into this framework~\cite{chen2023deep,chen2023evolving}. 

A second approach is the ``hardware as policy" paradigm, where the mechanical design itself is embedded into the control policy and jointly optimized with the parameters of the RL controller~\cite{chen2020hardware,sun2023co}. By representing physical parameters as differentiable components in a computational graph, gradient-based optimization algorithms can efficiently co-optimize neural network weights and mechanical design variables simultaneously~\cite{chen2020hardware,sun2023co,schaff2019jointly,luck2020data}. This enables more efficient and coordinated adaptation compared to bi-level methods, which treat design and control separately and often suffer from slow convergence. Such simultaneous optimization mirrors natural co-evolution, where animals adapt their body structures and behaviors together in response to growth and changing environments. While these methods have demonstrated success in enhancing adaptability, they often assume fixed training environments and struggle to incorporate real-world uncertainties. This potentially limits their robustness in practical applications.

The recent advancement of Digital Twin (DT) technology offers a transformative solution to CCD and other optimization problems~\cite{karkaria2025optimization}. By creating virtual representations of physical systems that evolve alongside their real-world counterparts, DTs facilitate continuous data assimilation and adaptive re-optimization~\cite{national2023foundational,thelen2022comprehensive,thelen2023comprehensive,zemskov2024security,karkaria2024towards,van2023digital, chen2025real}. This capability is particularly valuable for dynamic systems, such as vehicle suspension design, where real-time operational data can be leveraged to personalize suspension settings based on individual driving styles and preferences. Unlike traditional CCD approaches that rely on predefined models and assumptions, DTs provide a data-driven framework that continuously refines system models based on real-world conditions, making them more adaptable to varying operational demands.

Beyond real-time optimization, multi-generational design plays a crucial role in the evolution of engineering systems~\cite{van2023digital}. While real-time updates of digital models and control policies by DTs can enhance adaptability, re-optimization of physical components using accumulated data enables long-term design improvements, refining system configurations and material choices beyond immediate operational needs.  From one generation to the next, designers and engineers leverage data collected from the current generation to inform decision-making and guide system redesign~\cite{van2023digital}. This approach ensures that each successive iteration benefits from accumulated operational insights to improve performance, reliability, and efficiency. In the context of vehicle dynamics, DTs can simulate different road conditions, driving behaviors, and suspension configurations, allowing engineers to iteratively improve the design and operation of suspension systems~\cite{bhatti2021towards}. When the system is deployed in real-world conditions, RL controllers are continuously adapted and optimized based on online feedback data from sensors~\cite{schena2024reinforcement}.

% Reinforcement twinning (cite the paper) and other digital twins with RL, but none optimizes both physical systems and control policies

While previous works on RL-based DTs have demonstrated their effectiveness in optimizing system performance~\cite{xia2021digital} through the change of control policies, none have fully explored the simultaneous optimization of both physical system design and control policies. This paper addresses this gap by proposing a CCD framework for DT-enabled systems using Deep Reinforcement Learning (DRL). In addition, multi-generation design concepts are introduced, where data from previous generations are leveraged to improve both the accuracy of the digital model and system performance for future generations of physical systems. This research makes the following key contributions:
\begin{itemize}% Contributions
    % \item[\textbullet] We incorporate deep reinforcement learning to enable data-driven adaptability, allowing the system to respond effectively to changing and uncertain environments.
    \item [\textbullet] We extend CCD to multi-generational design following the DT framework that enables the physical system and control policy to co-adapt to dynamic and varying environments and improve overall performance over multiple generations of physical systems.
    \item [\textbullet] We propose a learning-based CCD method by integrating DRL with data-driven models to achieve rapid real-time decision-making and support iterative system redesign with model updates.
    % \item[\textbullet] We develop a framework that enables the physical system and control policy to co-adapt to dynamic and varying environments and improve overall performance. Automatic differentiation is used to efficiently compute gradients and integrate physical parameters into the learning process.
    % \item[\textbullet] By leveraging digital twins, the proposed framework facilitates multi-generational design optimization, supporting long-term product lifecycle management and iterative system redesign.
    \item[\textbullet] We incorporate uncertainty quantification (UQ) methods using physical data from previous generations to help designers make informed and robust decisions for next-generation designs.
    \item [\textbullet] We validate the proposed framework by demonstrating its application on an active suspension system, showcasing its ability to adapt to highly uncertain conditions while achieving improved performance through the co-optimization of physical components and control policies.
\end{itemize}

The remainder of this article is organized as follows: Section \ref{sec:Background} presents the preliminaries and technical background; Section \ref{sec:Method} presents the proposed CCD method and framework for DT-enabled systems with RL; Section \ref{sec:EngineeringStudy} implements and demonstrates the proposed approach with an engineering study of an active suspension system; and, Section \ref{sec:Conclusion} summarizes the conclusion and the future work.

\section{Technical Background}
\label{sec:Background}
\emph{Notation:} The sets of real numbers and non-negative integers are denoted by $\mathbb{R}$ and $\mathbb{N}_{\geq0}$, respectively. Given $a,b\in\mathbb{N}_{\geq0}$ such that $a<b$, we denote $\mathbb{N}_{[a,b]}:=\{a,a+1,...,b\}$. $\mathbf{x}_k$ denotes the state $\mathbf{x}$ at time $k$. The notation $\mathbf{I}_{a\times a}$ denotes an $a$-by-$a$ identity matrix. Given a random variable $X$, $\mathbb{E}[X]$ denotes its expected value. A Gaussian distribution with mean vector $\bm\mu$ and covariance matrix $\mathbf{\Sigma}$ is represented as $\mathcal{N}(\bm\mu,\mathbf{\Sigma})$.

\begin{figure}
    \centering
	\begin{subfigure}{0.49\textwidth}
        \centering
    	\includegraphics[width=0.55\textwidth]{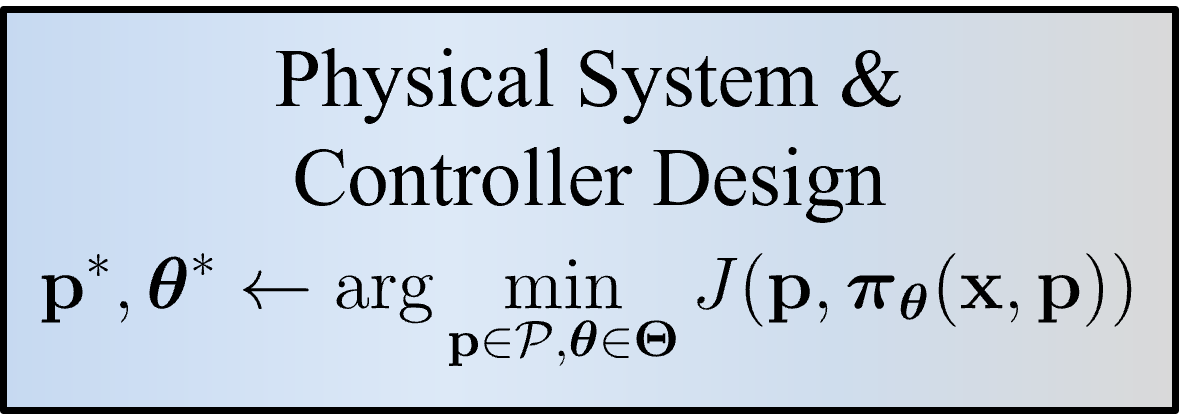}
    	\caption{}
    	\label{fig:SimCCD}
	\end{subfigure}
    \hfill
	\begin{subfigure}{0.49\textwidth}
        \centering
    	\includegraphics[width=0.55\textwidth]{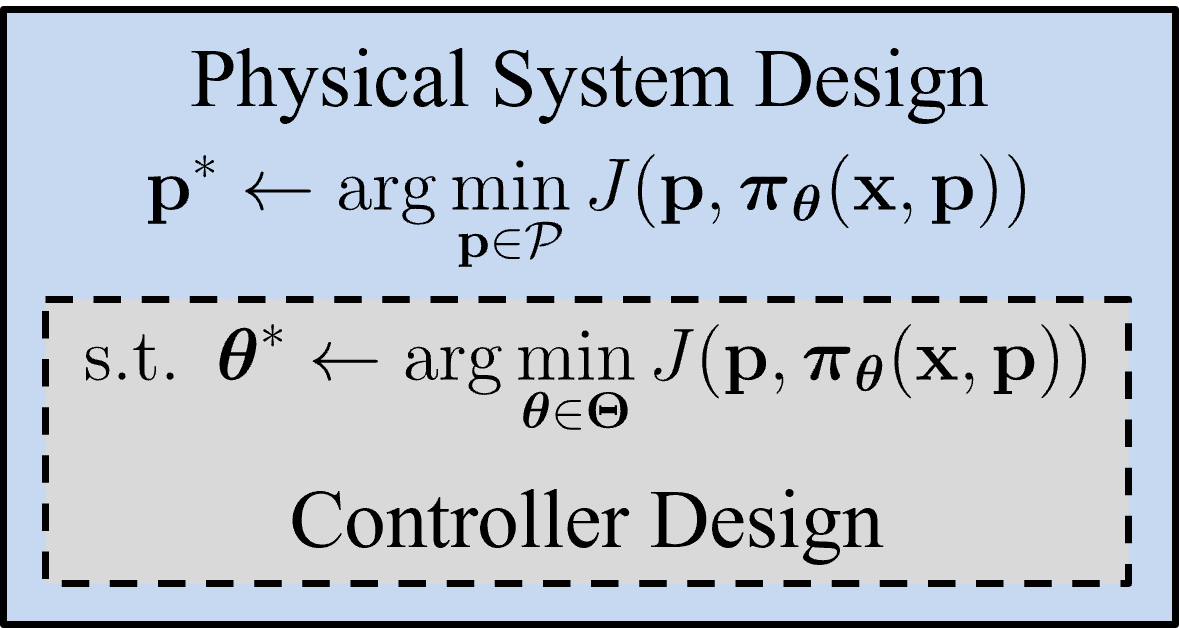}
    	\caption{}
    	\label{fig:NestedCCD}
	\end{subfigure}
    \caption{CCD formulations for feedback policy-based systems: (a) simultaneous and (b) nested (bi-level), where $\boldsymbol{\pi}_{\boldsymbol{\theta}}$ represents a feedback policy which is a function of states $\mathbf{x}$ and system parameters $\mathbf{p}\in\mathcal{P}$ with policy parameters $\boldsymbol{\theta}\in\mathbf{\Theta}$.}
    \label{fig:CCD_formulations}
\end{figure}

\subsection{Control Co-Design}
Control Co-Design (CCD) formulations integrate the physical system and controller design problems to explicitly account for the coupling between system dynamics and control strategies. This approach helps designers achieve system-level optimum compared to traditional sequential design methods~\cite{garcia2019control,allison2014special,tsai2022constraint}. There are two most common CCD formulations (simultaneous and nested)~\cite{herber2019nested}. Figure~\ref{fig:CCD_formulations} compares the simultaneous and nested formulations with a feedback control policy $\boldsymbol\pi_{\boldsymbol{\theta}}$ where $\boldsymbol{\theta}$ denotes the vector of policy parameters. Simultaneous CCD optimizes both physical and control design variables within a single optimization problem, while nested CCD separates physical and control design into an outer-loop (physical design) and an inner-loop (control optimization). Nested CCD is useful when the control problem requires specialized techniques or when different objective functions govern the physical and control system designs~\cite{nash2021robust,tsai2023robust,tsai2025control}.

The selection of CCD formulations depends on problem complexity, computational efficiency, and the availability of specialized control design tools. In this paper, we propose a CCD approach that uses the simultaneous CCD formulation to synergically optimize both physical system variables and control policies using learning-based techniques. The proposed method captures the intricate interactions between system design and control, allowing the system and the controller to co-adapt to dynamic and uncertain environments by learning from the real-time collected data.

\subsection{Reinforcement Learning}
\begin{figure}
    \centering
    \includegraphics[width=0.7\linewidth]{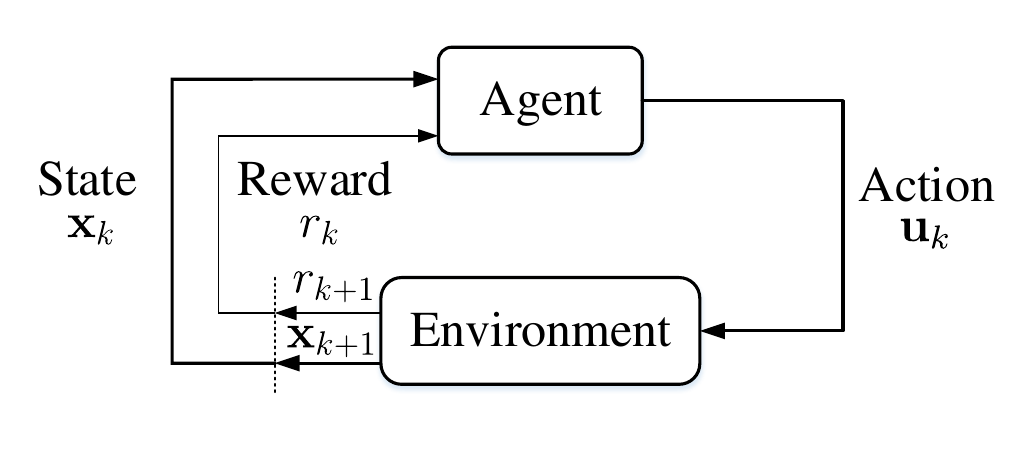}
    \caption{Diagram of reinforcement learning (RL), modified from \cite{sutton1998reinforcement}.}
    \label{fig:RL_diagram}
\end{figure}

% Basic + Value function and policy optimization
Reinforcement Learning (RL) is a framework for sequential decision-making where an agent interacts with an environment to learn an optimal policy that maximizes a cumulative reward~\cite{bucsoniu2018reinforcement}, shown as Fig.~\ref{fig:RL_diagram}. The RL problem is typically formulated as a Markov Decision Process (MDP), defined by a tuple $(\mathcal{X},\mathcal{U},P,R)$, where $\mathcal{X}$ is state space, $\mathcal{U}$ is action (control input) space, $P$ is transition function (dynamics) of a system, and $R$ is reward function (representing negative costs). In a MDP, the state transitions are typically stochastic, meaning that the next state $\mathbf{x}_{k+1}$ is not deterministic when an action $\mathbf{u}_k$ is applied to the system at state $\mathbf{x}_k$ at time $k\in\mathbb{N}_{\geq0}$. The transition function $P(\mathbf{x}_{k+1}|\mathbf{x}_k,\mathbf{u}_k)$ represents the probability of reaching state $\mathbf{x}_{k+1}$ from state $\mathbf{x}_k$ by taking action $\mathbf{u}_k$. However, in dynamics and controls, we prefer to represent the dynamic behavior by expressing the next state as a function of the current state and control action:
\begin{equation}
    \mathbf{x}_{k+1}=\mathbf{f}(\mathbf{x}_k,\mathbf{u}_k),
\end{equation}
which is a stochastic process according to the transition function $P(\mathbf{x}_{k+1}|\mathbf{x}_k,\mathbf{u}_k)$.

Once the system transits to $\mathbf{x}_{k+1}$, a scalar reward $r_{k+1}=R(\mathbf{x}_k,\mathbf{u}_k,\mathbf{x}_{k+1})$ is obtained, based on the reward function $R:\mathcal{X}\times \mathcal{U}\times \mathcal{X}\mapsto\mathbb{R}$. A policy is a mapping from states to probabilities of selecting each possible action. That is, the agent of RL chooses action $\mathbf{u}_k$ given state $\mathbf{x}_k$ is $\boldsymbol\pi(\mathbf{u}_k|\mathbf{x}_k)$. Given $(\mathbf{x}_0,\mathbf{u}_0,\mathbf{x}_1,\mathbf{u}_1,...)$ with the associated rewards $r_1,r_2,...,$, the return can be expressed as:
\begin{equation}
    G_k=\sum_{i=0}^{\infty}\gamma^i r_{k+i+1},
\end{equation}
where $\gamma\in(0,1]$ is the discount factor. The value function of state $\mathbf{x}$ under a policy $\boldsymbol\pi$, denoted as:
\begin{align}
    V^{\boldsymbol\pi}(\mathbf{x})=&\mathbb{E}_{\boldsymbol\pi}\left[G_k|\mathbf{x}_k=\mathbf{x}\right]\\\notag
    =&\mathbb{E}_{\boldsymbol\pi}\left[\sum_{i=0}^{\infty}\gamma^kR(\mathbf{x}_{k+i},\mathbf{u}_{k+i},\mathbf{x}_{k+i+1})\;\middle|\;\mathbf{x}_k=\mathbf{x}\right],
\end{align}
for all $\mathbf{x}\in \mathcal{X}$, represents the expected return when starting in state $\mathbf{x}$ and following $\boldsymbol\pi$ thereafter. We also call the function $V^{\boldsymbol\pi}(\mathbf{x})$ state-value function for policy $\boldsymbol\pi$ or V-function.

The state-value function $V^{\boldsymbol\pi}(\mathbf{x})$ only tells us the expected return from a state $\mathbf{x}$, assuming the policy $\boldsymbol\pi$ is followed. However, when making decisions, we need to evaluate specific actions rather than just states. We define the value of taking action $\mathbf{u}$ in state $\mathbf{x}$ under a policy $\boldsymbol\pi$, denoted:

\begin{align}  Q^{\boldsymbol\pi}(\mathbf{x},\mathbf{u})=&\mathbb{E}_{\pi}\left[G_k|\mathbf{x}_k=\mathbf{x},\mathbf{u}_k=\mathbf{u}\right]\notag\\
    =&\mathbb{E}_{\boldsymbol\pi}\left[\sum_{i=0}^{\infty}\gamma^kR(\mathbf{x}_{k+i},\mathbf{u}_{k+i},\mathbf{x}_{k+i+1})\;\middle|\;\mathbf{x}_k=\mathbf{x},\mathbf{u}_k=\mathbf{u}\right],
\end{align}
for all $\mathbf{x}\in \mathcal{X}$ and $\mathbf{u}\in \mathcal{U}$. We also call the function $Q^{\boldsymbol\pi}(\mathbf{x},\mathbf{u})$ action-value function for policy $\boldsymbol\pi$ or Q-function. $Q(\mathbf{x},\mathbf{u})$ provides this information by estimating the expected return for taking action $\mathbf{u}$ in state $\mathbf{x}$ and then following policy $\boldsymbol{\pi}$.

% DRL
Traditional RL methods, such as tabular Q-learning and dynamic programming, struggle when applied to complex control problems with continuous state and action spaces. In such cases, it becomes nearly impossible to explicitly store or update the value function for every possible state-action pair due to the infinite number of possibilities. This limitation makes traditional RL methods impractical for high-dimensional control problems, such as robotic manipulation, autonomous driving, or real-time optimization in DT-enabled systems.

Deep Reinforcement Learning (DRL) addresses this challenge by leveraging deep neural networks (DNNs) to approximate value functions, policies, or both~\cite{bucsoniu2018reinforcement,shakya2023reinforcement}. Instead of maintaining a lookup table for state-action values, DRL methods use neural networks as function approximators to generalize across similar states and actions. This enables RL algorithms to handle continuous, high-dimensional spaces efficiently. In particular, DRL has shown remarkable success in domains such as robotics, autonomous driving, and game playing, where large state-action spaces make traditional methods impractical. For example, policy-based methods, like Trust Region Policy Optimization (TRPO)~\cite{schulman2015trust} and Proximal Policy Optimization (PPO)~\cite{schulman2017proximal}, are widely used DRL algorithms because the algorithms exhibit more efficient and stable policy learning by incorporating trust region optimization. This ensures that policy updates remain within a constrained step size, thus preventing drastic changes that could destabilize learning. In addition to the aforementioned methods, numerous other DRL algorithms have been developed, each tailored to specific challenges such as sample efficiency, exploration, and stability. For a comprehensive review of DRL techniques, readers are referred to \cite{arulkumaran2017deep,ladosz2022exploration,wang2022deep}. By integrating deep learning, DRL significantly expands the applicability of RL to real-world problems where traditional methods would be computationally infeasible.

% Variants such as Asynchronous Advantage Actor-Critic (A3C)~\cite{mnih2016asynchronous}, Deterministic Policy Gradient Algorithms~\cite{silver2014deterministic}, and deep energy-based policies~\cite{haarnoja2017reinforcement} further extend the capabilities of DRL across different domains.

% \subsection{Co-optimization of Hardware and Policy with Reinforcement Learning}
% The integration of robot hardware design and control policy optimization, often referred to as co-design, has emerged as a critical area for enhancing robotic performance and adaptability~\cite{chen2021co}. Various computational frameworks have been proposed to tackle the co-design problem. A common formulation involves bi-level optimization, where low-level control policy learning is nested within a high-level tool or morphology design optimization~\cite{chen2023deep,sadat2023two}. To enhance the efficiency of high-level optimization, methods like Bayesian optimization and its multi-task variant have been employed as global optimization techniques~\cite{chen2023deep}. Chen \emph{et al.} integrate evolutionary computation with deep reinforcement learning and propose a co-adaption algorithm to simultaneously optimize the physical form (morphology) and its control mechanism~\cite{chen2023evolving}. These approaches aim to iteratively improve designs based on the performance of their corresponding control policies. 
\subsection{Co-Design of Robots with RL}

Given DRL’s success in learning complex control policies, it has been increasingly adopted in robotic co-design, where the goal is to jointly optimize both the physical morphology of robots and their control policies~\cite{ma2021diffaqua,yuhn20234d,chen2020hardware,sun2023co,schaff2019jointly,luck2020data}. This integrated perspective enables robots to evolve their mechanical structure and decision-making policies in tandem, yielding systems that can better adapt and perform in dynamic and uncertain environments~\cite{wang2023preco}.

A widely used strategy for solving CCD problems is to formulate them as a bi-level architecture. In this setup, the inner loop optimizes control policies for a fixed design candidate, while the outer loop updates the morphology through search methods such as evolutionary algorithms or Bayesian optimization~\cite{schaff2019jointly,wang2018neural,gupta2021embodied,chen2023c}. This hierarchical decomposition allows the use of tailored solvers for each layer, accommodating discrete or non-differentiable design spaces. However, bi-level methods are often computationally demanding and inefficient in sample usage, since every design modification requires retraining the controller from scratch~\cite{chen2023evolving}.

Since DRL has shown great potential in optimizing control policies, its application to robot co-design—where both the physical hardware and control policies for the robots are optimized simultaneously—has gained increasing attention~\cite{ma2021diffaqua,yuhn20234d,chen2020hardware,sun2023co,schaff2019jointly,luck2020data}. By extending the capabilities of DRL, robotic co-design focuses on developing algorithms that enable their mechanical structures to evolve and optimize along with their control strategies. Studies show that these robots have higher performance and adaptability in dynamic environments~\cite{wang2023preco}. A new paradigm within this field is the concept of ``hardware as policy," where the robot's mechanical design is considered analogous to a control policy and optimized jointly with its computational counterpart using DRL~\cite{chen2020hardware,sun2023co}. By modeling hardware as auto-differentiable computational graphs, gradient-based algorithms from policy optimization can be used to efficiently co-optimize mechanical parameters alongside neural network weights and biases~\cite{schaff2019jointly,luck2020data,ma2021diffaqua}.

\section{Method}
\label{sec:Method}
\begin{figure*}[b]
    \centering
    \includegraphics[width=\linewidth]{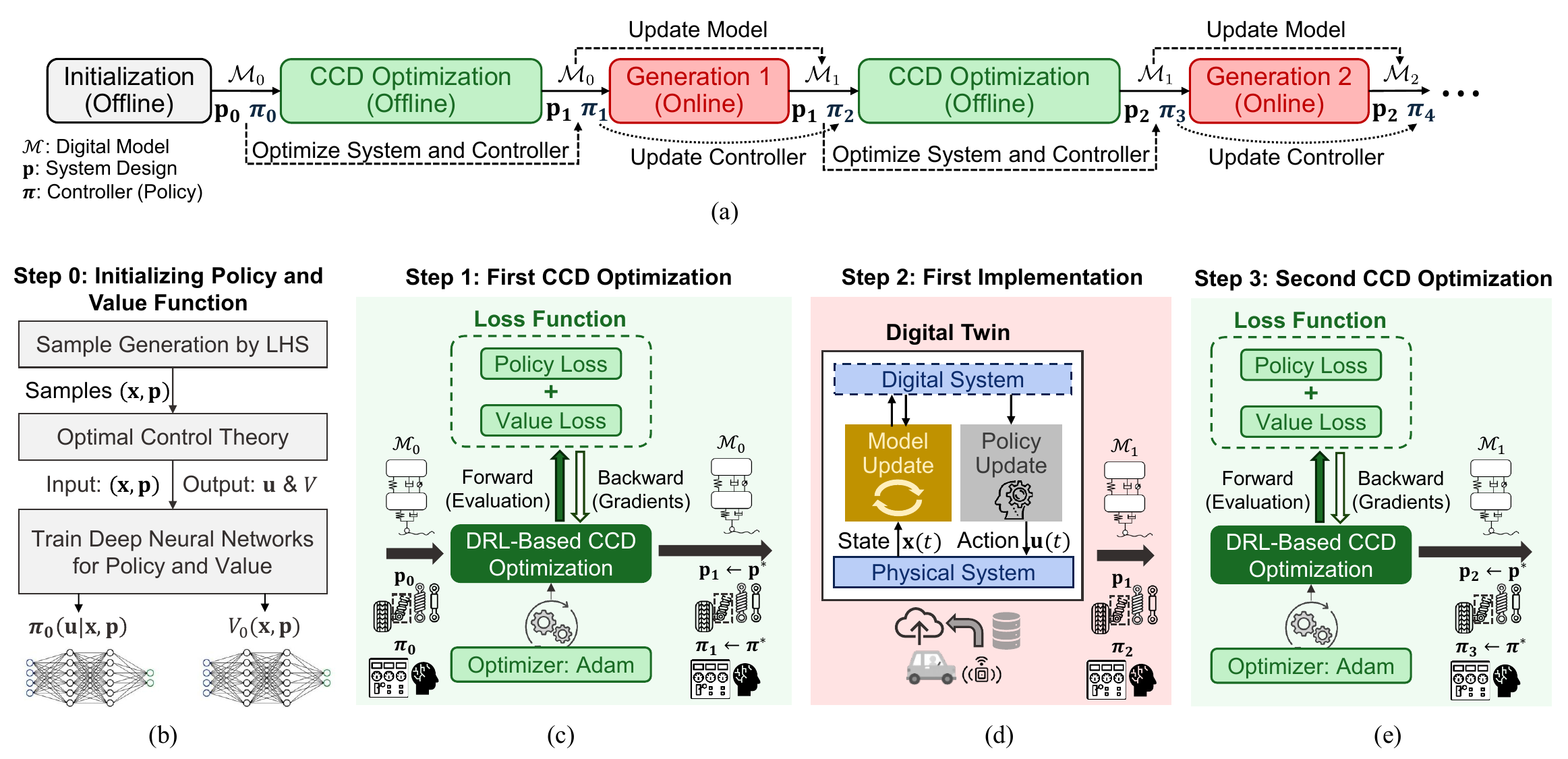}
    \caption{Proposed multi-generation Control Co-Design (CCD) framework for Digital Twin (DT) systems. (a) Overview of the CCD process across generations. (b) Offline initialization using Latin Hypercube Sampling (LHS) and optimal control theory to train initial policy $\boldsymbol{\pi}_0$ and value function $V_0$. (c) First CCD optimization using gradient-based DRL with auto-differentiation. (d) Online deployment and data collection in Generation-1, where real-time feedback is used to update the digital model and policy. (e) Second CCD optimization using updated models to improve performance in subsequent generations.}
    \label{fig:DTCCD}
\end{figure*}

Traditional CCD methods rely on predefined mathematical models, limiting their adaptability to real-time variability in dynamic environments. This lack of adaptability leads to performance degradation in applications like vehicles and wind turbines due to wear and environmental changes~\cite{van2023digital}. To overcome this, the proposed CCD method uses DT frameworks for real-time data collection, continuous model updates, and informed decision-making on system design and policy updating. By integrating operational data, UQ, and DRL, our approach can iteratively refine the system model and improve designs for subsequent generations. This integration enhances the performance, adaptability, and robustness of these systems, which are difficult to achieve with traditional CCD methods.

\subsection{Overview of Proposed Framework}

This paper introduces a CCD framework for DTs that leverages DRL within a multi-generation design paradigm. The proposed framework enables the DT to evolve iteratively through successive learning cycles by continuously integrating data collected from physical systems during operation. Across generations, operational data is used to refine the digital model, incorporating quantile-based learning to account for environmental uncertainties and policy learning to improve decision-making in dynamic and uncertain conditions. Prior to advancing to the next generation, the updated digital model, which is better aligned with real-world behaviors and variability, is employed to re-optimize the CCD problem. This iterative refinement ensures that both system designs and control strategies become progressively more adaptive and efficient over time.

Figure~\ref{fig:DTCCD} illustrates the conceptual workflow of the proposed framework, composed of four steps. Prior to \textbf{Step 1}, an initialization stage is performed to construct the initial policy $\boldsymbol{\pi}$ and value function $V_0$ using data generated by optimal control theory and Latin Hypercube Sampling (LHS), which provides a crucial foundation for guiding early learning and improving sample efficiency, shown in Fig.~\ref{fig:DTCCD}(b). In \textbf{Step 1}, shown in Fig.~\ref{fig:DTCCD}(c), the first CCD optimization is performed using the virtual model $\mathcal{M}_0$, initial physical design $\mathbf{p}_0$, and initial control policy $\boldsymbol{\pi}_0$. Solving this problem yields the optimal system design $\mathbf{p}_1$ and control policy $\boldsymbol{\pi}_1$, which are deployed in the real environment in \textbf{Step 2} (referred to as Generation 1), shown in Fig.~\ref{fig:DTCCD}(d). During this phase, physical data is collected, and the digital model and the DRL-based control policy are updated in real time, resulting in an updated model $\mathcal{M}_1$ and an updated controller $\boldsymbol{\pi}_2$, respectively.

In \textbf{Step 3}, shown in Fig.~\ref{fig:DTCCD}(e), the refined model $\mathcal{M}_1$ is used to re-solve the CCD optimization problem, generating a new system design $\mathbf{p}_2$ and updated controller $\boldsymbol{\pi}_2$. This is followed by \textbf{Step 4}, where the new design and policy are implemented in the physical environment, initiating another cycle of data collection and model refinement, leading to $\mathcal{M}_2$. This process continues iteratively across generations, enabling continuous adaptation and performance improvement of both the system and its controller.

\subsection{Illustrative Example}
\label{sec:IllustrativeExample}

To demonstrate the proposed framework and its individual steps, we consider a simplified illustrative example. The system dynamics are described by a discrete-time linear model:

\begin{equation}
    \mathbf{x}_{k+1}=
    \underbrace{\begin{bmatrix}
    0.8~&~0.5\\0.5~&~0.6
    \end{bmatrix}}_{\mathbf{A}}
    \mathbf{x}_k+
    \underbrace{\begin{bmatrix}
    0.5\\p
    \end{bmatrix}}_{\mathbf{B}}
    u_k+\mathbf{w}_k,
    \label{eq:numerical_example}
\end{equation}
where $\mathbf{x}_k$ denotes the state vector at time step $k$ constrained by $\mathcal{X}:=\left\{\mathbf{x}\in\mathbb{R}^2~|~x_1\in[-10,5], x_2\in[-5,2]\right\}$, and $u_k \in \mathcal{U} := \{ u \in \mathbb{R} \mid -1 \leq u \leq 1 \}$ is the bounded control input. The term $\mathbf{w}_k \in \mathbb{R}^2$ represents process noise or disturbances at time $k$, and the scalar parameter $p$ serves as a tunable physical design variable within the input matrix $\mathbf{B}$.

The parameter $p$ plays a critical role in the system's controllability. While increasing $p$ enhances the system's responsiveness to control inputs, it may also amplify the sensitivity to control-induced errors, thus introducing a trade-off in the design process.

The objective is to regulate the system to the origin. To this end, we define the reward function as:

\begin{equation}
    r_{k+1}=-\mathbf{x}_{k+1}^\top\mathbf{Q}\mathbf{x}_{k+1}-0.1u_k^2,
    \label{eq:reward_numerical}
\end{equation}
where $\mathbf{Q} = \mathbf{I}_{2\times2}$ is the identity matrix used to penalize deviations from the origin in the state space, and the second term penalizes control effort.

Prior to system deployment, no physical data or prior environmental knowledge is assumed. To simulate real-world uncertainty, the disturbance $\mathbf{w}_k$ is modeled as a zero-mean multivariate Gaussian noise process, given by: 
\begin{equation}
    \mathbf{w}_k\sim\mathcal{N}\left(\begin{bmatrix}0\\0\end{bmatrix},\begin{bmatrix}0.1^2&0\\0&0.2^2\end{bmatrix}\right),~\forall k\in\mathbb{N}_{\geq0}.
    \label{eq:disturbance_numerical}
\end{equation}

It is noted that the system in Eq.~(\ref{eq:numerical_example}) is inherently unstable since $||\mathbf{A}^k||\to\boldsymbol\infty$ as $k\to\infty$. This instability highlights the necessity of an effective control strategy in conjunction with a well-chosen physical design $p$ to stabilize the system, emphasizing the importance of proper policy initialization prior to learning.

\subsection{Step 0: Initialization}

A key challenge in DRL-based CCD lies in the initialization of the policy and value function approximators, typically represented by deep neural networks. Traditional methods often require solving multiple optimal control problems for diverse initial conditions, which is computationally demanding and may not scale well to high-dimensional design spaces. Poor initialization can also lead to unstable training and slower convergence, which reduces the practicality of DRL in real-time engineering applications.

To address this challenge, we adopt a sample generation strategy inspired by the set invariance and explicit MPC framework proposed by Chen et al.~\cite{chen2018approximating}. Their approach enables efficient sampling of feasible state-action pairs that satisfy system constraints and approximate optimality, without the need to repeatedly solve optimization problems. This provides a data-driven yet control-theoretically grounded way to initialize neural network-based controllers.

\subsubsection{Incorporating Physical Design Parameters}
Inspired by \cite{schaff2019jointly,luck2020data,chen2020hardware,tsai2023control}, the input space extends beyond system states to include physical design variables $\mathbf{p}$ in our CCD setting. To accommodate this, we generalize the method of Chen et al.~\cite{chen2018approximating} by introducing physical parameters $\mathbf{p}$ directly into the input of the policy and value functions, yielding $\boldsymbol{\pi}(\mathbf{u|x,p})$ and $V(\mathbf{x}, \mathbf{p})$. For the illustrative example, since the physical parameter is a scalar $p$, we only add one more dimension to the sampling space. We will generate triplets $(\mathbf{x},{p}, {u})$ where ${u}$ is obtained by following Section \ref{sec:FeasibleSampleGeneration}. These samples provide a structured basis for initializing the neural networks, improving stability and constraint satisfaction in early training phases.

\subsubsection{Feasible Sample Generation via Invariant Set Computation}
\label{sec:FeasibleSampleGeneration}
\begin{figure}
    \centering
	\begin{subfigure}{0.24\textwidth}
        \centering
    	\includegraphics[width=\textwidth]{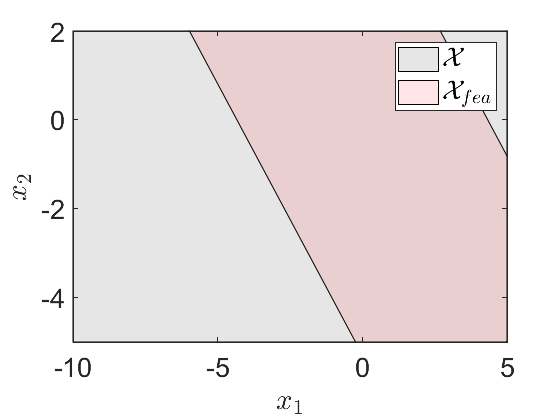}
    	\caption{$p=0.5$}
    	\label{fig:Xfea1}
	\end{subfigure}
    \hfill
	\begin{subfigure}{0.24\textwidth}
        \centering
    	\includegraphics[width=\textwidth]{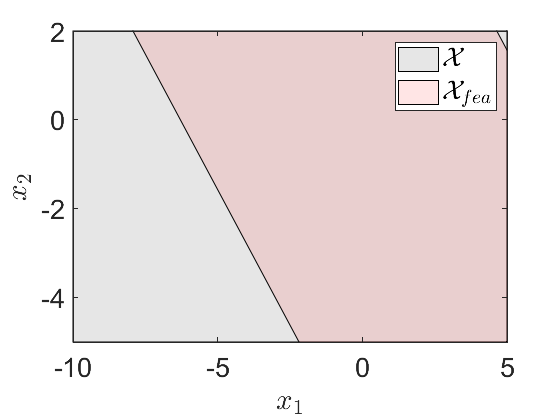}
    	\caption{$p=1.0$}
    	\label{fig:Xfea2}
	\end{subfigure}
    \begin{subfigure}{0.24\textwidth}
        \centering
    	\includegraphics[width=\textwidth]{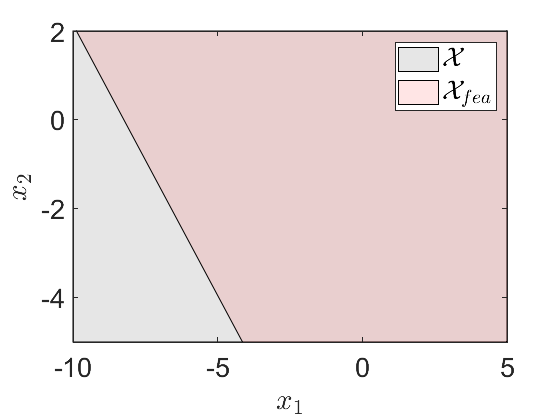}
    	\caption{$p=1.5$}
    	\label{fig:Xfea3}
	\end{subfigure}
    \hfill
	\begin{subfigure}{0.24\textwidth}
        \centering
    	\includegraphics[width=\textwidth]{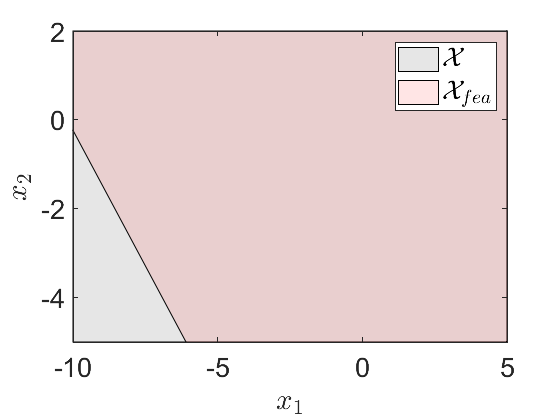}
    	\caption{$p=2.0$}
    	\label{fig:Xfea4}
	\end{subfigure}
    \caption{Visualization of $\mathcal{X}_{fea}$ of the illustrative example for different values of $p$. The pink area in each plot represents the state feasible region $\mathcal{X}_{fea}$ with the specified value of $p$, which expands as $p$ increases. This trend reflects improved controllability of the system due to the larger entries in Matrix $\mathbf{B}$, allowing a broader set of states to be driven to the origin under the state and control input constraints.}
    \label{fig:Xfea}
\end{figure}

To ensure that all generated samples are within regions of safe operation and satisfy both state and input constraints, the sample generation procedure is introduced and detailed in the following steps:
\begin{enumerate}
    \item Define the discrete-time linear system $\mathbf{x}_{k+1} = \mathbf{A} \mathbf{x}_k + \mathbf{B} \mathbf{u}_k$, where $\mathbf{A}$, $\mathbf{B}$, or both depend on the physical parameters $\mathbf{p}$. For the illustrative example in Eq.~(\ref{eq:numerical_example}), only $\mathbf{B}$ depends on the physical parameter ${p}$.
    \item Sample a set of candidate points $(\mathbf{x}, \mathbf{p})$ uniformly within predefined bounds using LHS.
    \item For each sample, compute its feasible $\mathcal{X}_{{fea}}(\mathbf{p})$ using tools such as \texttt{MPT3}~\cite{MPT3}. Figure \ref{fig:Xfea} visualizes $\mathcal{X}_{{fea}}(p)$ for the illustrative example (Eq.~(\ref{eq:numerical_example})) with $p=0.5$, $1.0$, $1.5$, and $2.0$.
    \item Accept the sample $(\mathbf{x}, \mathbf{p})$ only if $\mathbf{x} \in \mathcal{X}_{{fea}}(\mathbf{p})$.
    \item For each accepted pair, assign a control input $\mathbf{u}$ using the corresponding explicit MPC policy (detailed in \cite{chen2018approximating}).
\end{enumerate}

For the illustrative example (Eq.~(\ref{eq:numerical_example})), 300 samples $(\mathbf{x}, {p}, {u})$ are generated and then used to pretrain the deep neural networks (DNNs) that approximate the policy and value functions. Both networks are implemented as fully connected feedforward architectures with four hidden layers. Each network takes a 3-dimensional input vector: comprising the system state $\mathbf{x} = \left[x_1, x_2\right]^\top$ and the scalar physical design parameter $p$. The hidden layers are configured with 32, 32, 16, and 16 neurons, respectively, with the first layer using a \texttt{Tanh} activation function to facilitate early nonlinear learning. The value network outputs a scalar estimate of the expected return, while the policy network is split into two branches: one for predicting the mean control action and the other for the standard deviation of the Gaussian policy. The mean network is pretrained using the sampled control actions from explicit MPC, while the standard deviation branch is initialized with small constant bias values (e.g., 0.01) and zero weights to allow for controlled exploration at the start of training.

This initialization technique significantly reduces the reliance on online optimal control solvers for pretraining and provides high-quality, constraint-satisfying data. By incorporating physical design variables directly into the input space, the method ensures that both structural and control-related decisions are grounded in feasible operation, enabling smoother and more sample-efficient DRL training.

\begin{figure}[t]
    \centering
    \includegraphics[width=0.95\linewidth]{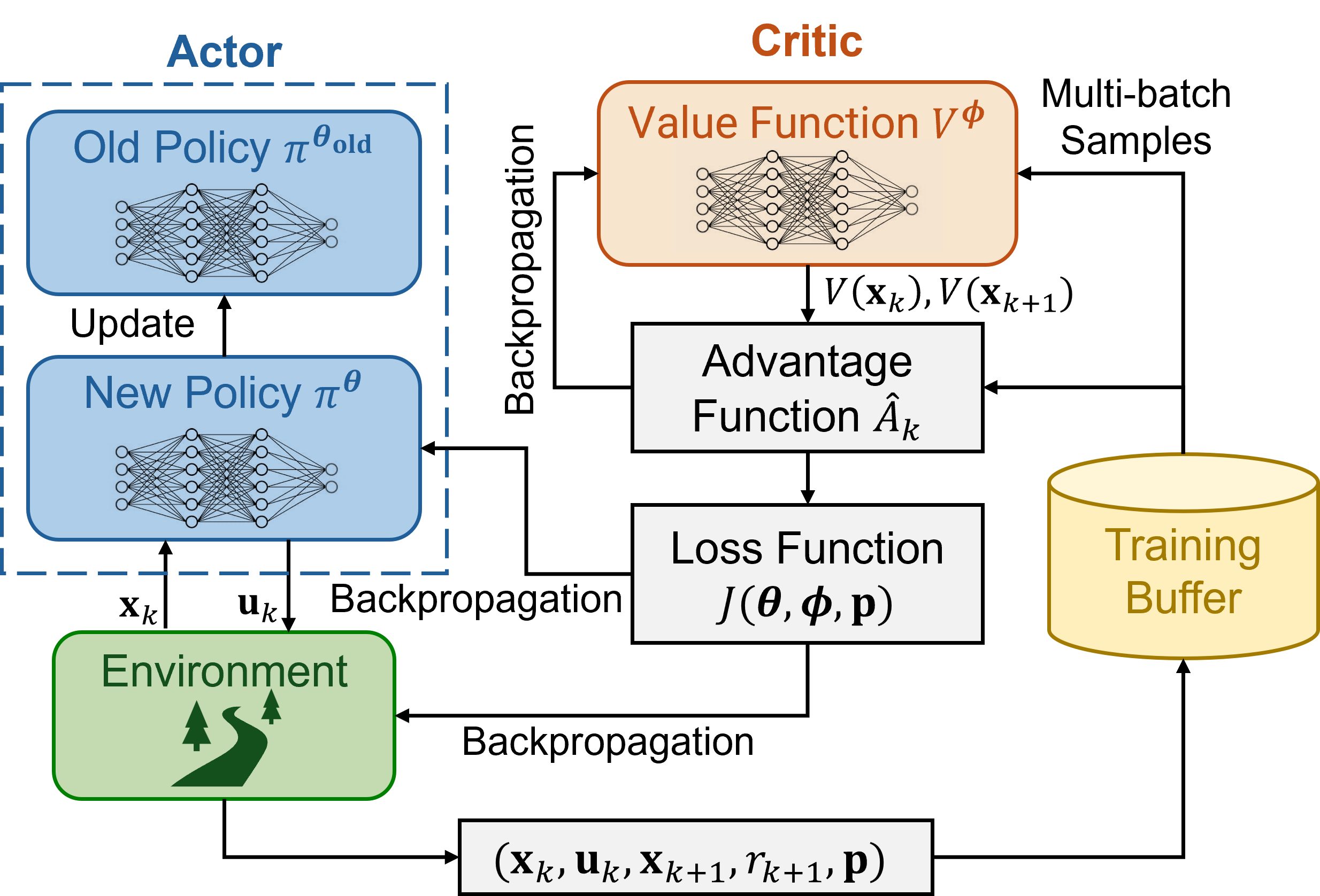}
    \caption{Flow chart of the proposed DRL-based CCD optimization using the PPO algorithm. Unlike conventional PPO, this framework integrates the physical system parameter $\mathbf{p}$ as part of the input space and enables gradient-based updates of both the control policy and the physical design. This is achieved by treating the environment dynamics as differentiable and enabling backpropagation through the environment to the loss function.}
    \label{fig:PPO_CCD}
\end{figure}

\subsection{Step 1: DRL-Based CCD Optimization}
\label{sec:CCD_RL}
Following the initialization of the neural network approximators, we present a DRL framework based on Proximal Policy Optimization (PPO) to solve the CCD problem, shown in Fig.~\ref{fig:PPO_CCD}. Unlike traditional PPO implementations, which focus solely on optimizing policy and value network parameters, our method simultaneously updates the physical design variables $\mathbf{p}$ along with the policy parameters $\boldsymbol{\theta}$ and value function parameters $\boldsymbol{\phi}$ during training. This tight coupling allows both the physical system and the control policy to co-adapt through reinforcement learning.

\begin{figure}[t]
    \centering
	\begin{subfigure}{0.49\textwidth}
    	\hspace{0.4cm}\includegraphics[width=0.85\textwidth]{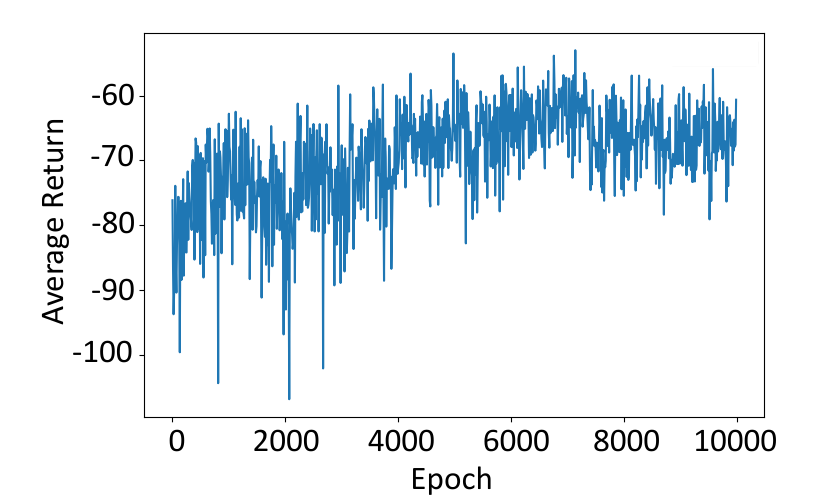}
    	\caption{}
    	\label{fig:return_history_numerical}
	\end{subfigure}
    \hfill
	\begin{subfigure}{0.49\textwidth}
    	\hspace{0.4cm}\includegraphics[width=0.85\textwidth]{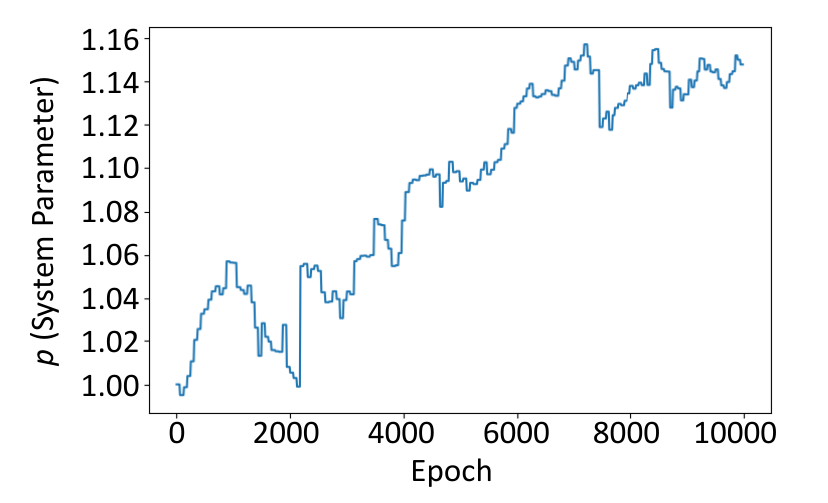}
    	\caption{}
    	\label{fig:p_history_numerical}
	\end{subfigure}
    \caption{Training history for numerical example with (a) average return and (b) system parameter $p$.}
    \label{fig:history_numerical}
\end{figure}

\begin{figure}
    \centering
    \includegraphics[width=0.8\linewidth]{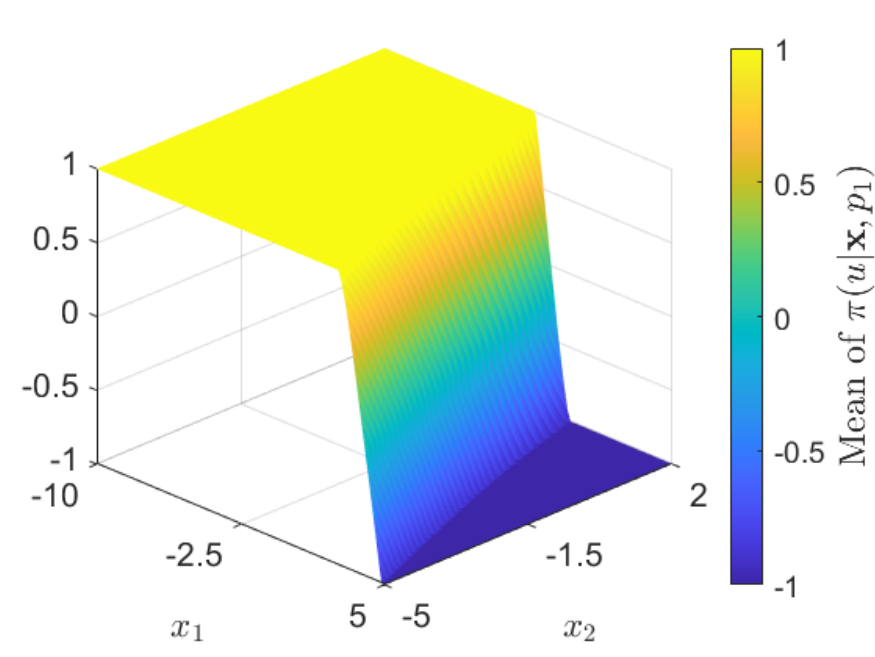}
    \caption{Visualization of the mean of the optimal policy $\pi(u|\mathbf{x},p_1)$, where $p_1=1.1513$.}
    \label{fig:policy1_numerical}
\end{figure}

To achieve this, we formulate a joint optimization problem that minimizes a loss function composed of a clipped surrogate policy objective and a value prediction error:

\begin{align}
   \min_{\boldsymbol\theta, \boldsymbol\phi, \mathbf{p}}\; &J(\boldsymbol\theta, \boldsymbol\phi, \mathbf{p})\notag\\
   =\mathbb{E}_k &\left[
\underbrace{-\min \left( \rho_k(\boldsymbol\theta, \mathbf{p}) \hat{A}_k,L^ \text{CLIP}(\rho_k(\boldsymbol\theta, \mathbf{p}), 1-\epsilon, 1+\epsilon)\hat{A}_k \right)}_{\text{Policy Loss}}\right.\notag\\
&\left.+ c_v \cdot \underbrace{L_{\text{SmoothL1}}\left(V^{\boldsymbol\phi}(\mathbf{x}_k, \mathbf{p}),\hat{V}_k\right)}_{\text{Value Loss}}\right],
\label{eq:PPO_loss}
\end{align}
where $\hat{A}_k$ is the advantage function at time step $k$, which estimates how much better an action is compared to the expected value of a state, and $L^ \text{CLIP}(\rho_k(\boldsymbol{\theta},\mathbf{p}),1-\epsilon,1+\epsilon)$ is a function clipping $\rho_k(\boldsymbol{\theta},\mathbf{p})$ within $(1-\epsilon,1+\epsilon)$, $\rho_k$ denotes the probability ratio between the new and old policies:
\begin{equation}
    \rho_k(\boldsymbol{\theta},\mathbf{p})=\frac{\boldsymbol{\pi}^{\boldsymbol\theta}\left(\mathbf{u}_k|\mathbf{x}_k,\mathbf{p}\right)}{\boldsymbol{\pi}^{\boldsymbol\theta_{\text{old}}}\left(\mathbf{u}_k|\mathbf{x}_k,\mathbf{p}_{\text{old}}\right)},
\end{equation}
$c_v$ is the coefficient for value loss, and we use Smooth L1 loss to define the value loss:
\begin{equation}
    L_{\text{SmoothL1}}(a, b) = 
    \begin{cases}
        \frac{1}{2}(a - b)^2, & \text{if } |x - y| < 1 \\
        |a - b| - \frac{1}{2}, & \text{otherwise.}
    \end{cases}
\end{equation}

The inclusion of $\mathbf{p}$ in both policy and value networks enables the optimizer to compute gradients not only with respect to the network weights but also with respect to the physical design parameters. To facilitate efficient gradient computation, we utilize automatic differentiation via the \texttt{PyTorch} autograd engine~\cite{paszke2019pytorch}, which builds the computational graph and performs backpropagation for all trainable parameters, including $\mathbf{p}$. This allows for joint, end-to-end updates during the learning process, enhancing sample efficiency and enabling exploration across both design and control spaces.

Figure~\ref{fig:history_numerical} presents the training histories of the average returns and the system parameter $p$ for the numerical example introduced in Section~\ref{sec:IllustrativeExample}. The DRL-based CCD optimization was run for 10,000 epochs using a learning rate of $10^{-5}$. The training process required approximately 7 minutes on a machine with an AMD EPYC 7413 24-Core CPU (2.65 GHz), 1 TB RAM, and a 64-bit Linux operating system. As shown in the figure, the system parameter $p$ evolves over the course of training, reflecting the co-adaptation of the physical system and its control policy. Specifically, Fig.~\ref{fig:p_history_numerical} illustrates how $p$ initially increases to enhance system controllability. However, as excessively large values of $p$ can amplify sensitivity to control inputs—especially if the policy is not yet fully trained—the learning algorithm occasionally reduces $p$ to balance performance gains with robustness.

Figure~\ref{fig:policy1_numerical} visualizes the mean of the learned policy $\pi(u|\mathbf{x},p_1)$ for the final optimized system parameter $p_1 = 1.1513$. Since the policy is stochastic by design, the control action may vary upon each execution. For visualization, we plot the mean of the distribution; in deployment, however, the policy continues to sample actions probabilistically.

\begin{figure*}
    \centering
    \includegraphics[width=0.8\linewidth]{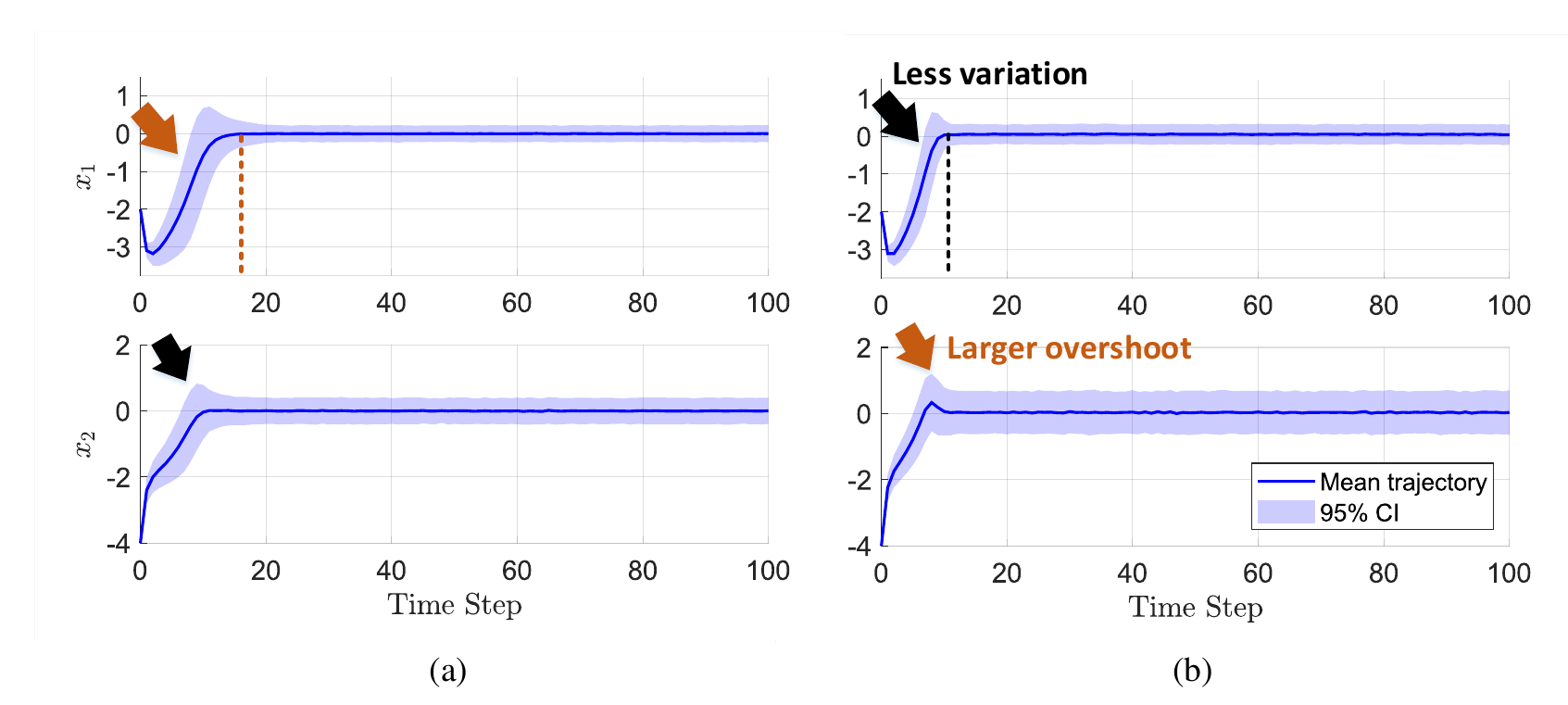}
    \caption{State trajectories (a) before and (b) after the first CCD optimization for the numerical example with 1000 replicates. The shaded regions show the 95\% confidence intervals (CIs).}
    \label{fig:traj_before_after_CCD_numerical}
\end{figure*}

To evaluate the performance of the optimized design, we simulate 1000 independent trajectories. The resulting state responses exhibit reduced settling times and smaller variations in overshoot compared to those before optimization, shown in Fig.~\ref{fig:traj_before_after_CCD_numerical}. Quantitatively, the average return improves from -86.328 (pre-optimization) to -69.783 (post-optimization), while the standard deviation decreases from 26.369 to 11.994. These results highlight that the proposed DRL-based CCD method successfully enhances dynamic performance by jointly optimizing the physical system parameter and control policy.

\subsection{Step 2: First Implementation (Generation 1)}
\label{sec:Step1}

Prior to deploying the system in Generation 1, it is important to recognize the inherent mismatch between the digital model and the actual physical system. These discrepancies stem from unmodeled environmental variations, unknown physical dynamics, and nonlinear effects that the nominal model does not fully capture. For instance, real-world mechanical components such as springs and dampers often exhibit nonlinear behaviors, deviating from the idealized linear assumptions. Furthermore, external disturbances, such as variable road conditions and fluctuating vehicle speeds, introduce uncertainties that can only be observed during physical implementation.

To reflect these modeling discrepancies, we construct a hypothetical physical system by augmenting the nominal model with additional disturbance and nonlinear terms:

\begin{align}
    \mathbf{x}_{k+1}&=
    \mathbf{A}
    \mathbf{x}_k+
   \mathbf{B}
    u_k+\mathbf{w}_k+\underbrace{\begin{bmatrix}0.1\\-0.2\end{bmatrix}+\text{U}\left(\begin{bmatrix}-0.1\\-0.1\end{bmatrix},\begin{bmatrix}0.1\\0.1\end{bmatrix}\right)}_{\text{Unknown Disturbances}}\notag\\
    &+\underbrace{\left(\begin{bmatrix}0.1&0\\0&0.1\end{bmatrix}\text{diag}\left(\sin\mathbf{x}_k\right)+\begin{bmatrix}0&0.1\\0.1&0\end{bmatrix}\text{diag}\left(\cos\mathbf{x}_k\right)\right)\mathbf{x}_k}_{\text{Unknown Nonlinearity}}\notag\\
    &+\underbrace{\left(\begin{bmatrix}0.5&0\\0&0\end{bmatrix}\sin\mathbf{x}_k+\begin{bmatrix}0&0\\0&0.5p\end{bmatrix}\cos\mathbf{x}_k\right)u_k}_{\text{Unknown Nonlinearity}},
    \label{eq:real_system_numerical_example}
\end{align}
where $\text{U}(lb, ub)$ represents a uniform distribution between the lower bound $lb$ and upper bound $ub$. These additional terms simulate real-world uncertainties (such as additive disturbances and input/state-dependent nonlinearities) that are not incorporated in the original digital model defined in Eq.~(\ref{eq:numerical_example}). This modeling gap emphasizes the importance of using a digital twin framework that can iteratively adapt based on physical observations collected during operation.

Despite the presence of discrepancies between the digital model and the real physical system, the DRL-based controller remains highly effective in maintaining control performance. This robustness stems from the inherently adaptive nature of learning algorithms, which are specifically designed to learn and evolve in dynamic and uncertain environments. By incorporating real-time data collected from the physical system, the RL agent continuously refines its policy, enabling it to make more informed and proactive decisions that maximize long-term rewards. As a result, the controller can adapt to variations in system behavior, thereby mitigating the impact of modeling inaccuracies and unanticipated disturbances.

To explicitly account for the mismatch between the digital and physical systems, we employ a discrepancy model trained on physical data collected during deployment. This model captures deviations by leveraging UQ techniques. Specifically, quantile regression, a data-driven method for estimating conditional quantiles of a response variable. Quantile regression is well-suited for modeling uncertainties in dynamic systems, as it provides predictive intervals that characterize both systematic bias and aleatoric uncertainty (i.e., uncertainty inherent in the data)~\cite{chen2025uncertainty}. It also provides conformal estimation of the distribution which is free from assuming the distribution to be Gaussian, providing additional learning flexibility. 

By selecting representative quantile levels, such as the 10th, 50th (median), and 90th percentiles, the discrepancy model captures the spread and central tendency of prediction errors. In the context of the numerical example, the model estimates the error at the next time step as a function of the current state, control input, and current error:
\begin{equation}
   \left[\mathbf{e}^{\text{upper}}_{k+1}, \mathbf{e}^{\text{median}}_{k+1},\mathbf{e}^{\text{lower}}_{k+1}
    \right]^\top=\mathbf{f_e}(\mathbf{e}_k,\mathbf{x}_k,u_k),
    \label{eq:quantile_numerical}
\end{equation}
where $\mathbf{e}_k := \mathbf{x}_k - \bar{\mathbf{x}}_k$ denotes the deviation between the actual state $\mathbf{x}_k$ and the nominal predicted state $\bar{\mathbf{x}}_k$ from the digital model, computed using $\bar{\mathbf{x}}_{k+1} = \mathbf{A} \bar{\mathbf{x}}_k + \mathbf{B} u_k$. The predicted quantiles $\mathbf{e}^{\text{upper}}_{k+1}$, $\mathbf{e}^{\text{median}}_{k+1}$, and $\mathbf{e}^{\text{lower}}_{k+1}$ represent the range of likely deviations in the next state.

\begin{figure}[t]
    \centering
    \includegraphics[width=0.9\linewidth]{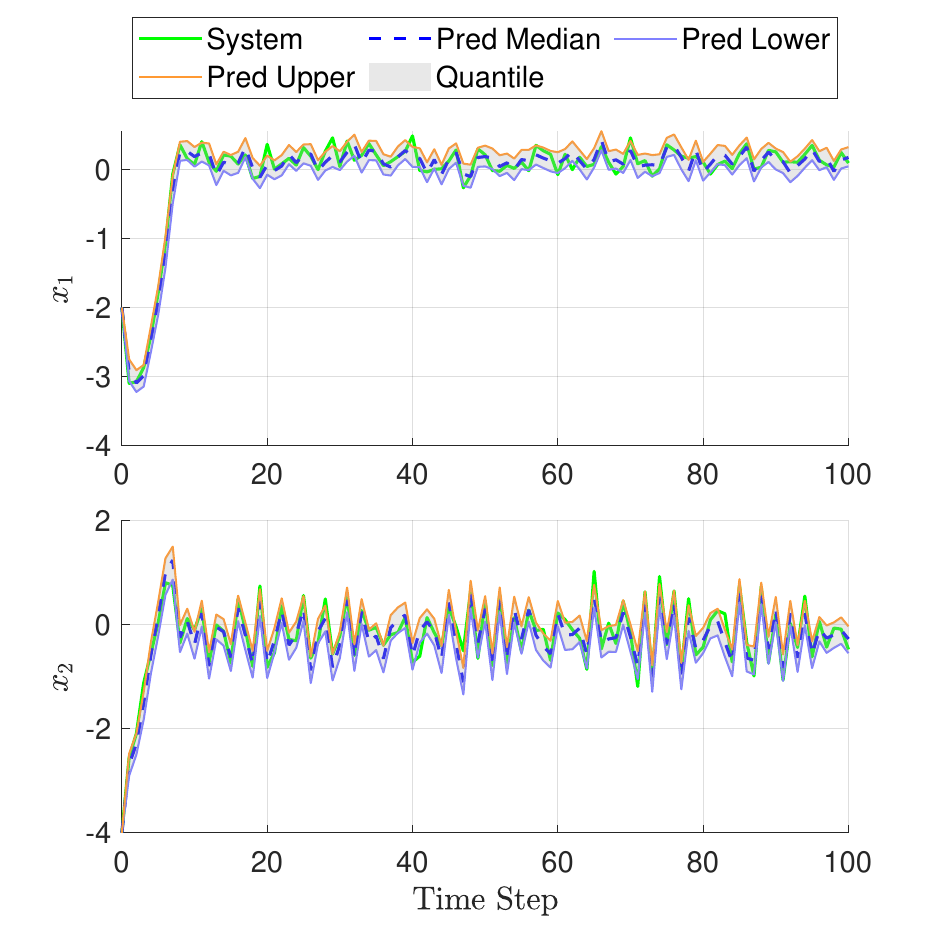}
    \caption{Visualization of the learned quantiles and the real system trajectories for the numerical example.}
    \label{fig:quantile_numerical}
\end{figure}

In this formulation, we assume that the primary sources of uncertainty stem from the external environment, rather than variability in the physical system parameter $p$. Therefore, the discrepancy model is designed to be independent of $p$. Figure~\ref{fig:quantile_numerical} illustrates the predicted quantiles (10th, 50th, and 90th percentiles), capturing the uncertainty bounds of the model prediction. The model achieves a root mean square error (RMSE) of 0.1355 on the validation dataset, indicating reliable accuracy in capturing the discrepancy between the real and nominal systems.

\subsection{Step 3: Second CCD Optimization (Generation 2)}

Instead of continuing with the nominal digital model subject to stochastic disturbances as defined in Eq.~(\ref{eq:numerical_example}), we refine the model by incorporating the learned discrepancy function $\mathbf{f_e}$ obtained in \textbf{Step 2}. The updated dynamics are given by:
\begin{equation}
    \mathbf{x}_{k+1}=
    \mathbf{A}
    \mathbf{x}_k+
   \mathbf{B}
    u_k+\mathbf{e}^{\text{median}}_{k+1},
    \label{eq:updated_dynamic_model}
\end{equation}
where $\mathbf{e}^{\text{median}}_{k+1}$ denotes the median predicted quantile of the error, as estimated by the discrepancy model in Eq.~(\ref{eq:quantile_numerical}). To reflect this model refinement in the learning process, the reward function is also updated as follows:
\begin{equation}
    r_{k+1}=-\mathbf{x}_{k+1}^\top\mathbf{Q}\mathbf{x}_{k+1}-0.1u_k^2-\Delta\mathbf{e}_{k+1}^\top\mathbf{Q}_q\Delta\mathbf{e}_{k+1},
    \label{eq:updated_reward_numerical}
\end{equation}
where $\Delta\mathbf{e}_{k+1} := \mathbf{e}^{\text{upper}}_{k+1} - \mathbf{e}^{\text{lower}}_{k+1}$ captures the predicted uncertainty range (i.e., quantile width) at time step $k+1$, and $\mathbf{Q}_q = 0.1 \mathbf{Q}$ is a weighting matrix applied to penalize uncertainty. This revised reward function encourages the RL agent not only to maximize performance but also to minimize predictive uncertainty, thereby promoting more robust design and control decisions.

In \textbf{Step 3}, we perform a second CCD optimization using the updated digital model and reward formulation. The new optimized physical design parameter is $p_2 = 1.2947$. As previously discussed, increasing $p$ improves controllability but requires a sufficiently capable controller. The upward shift in $p$ reflects the co-adaptation of the physical system and the RL policy, indicating that the controller has matured in its ability to manage a more sensitive system.

To evaluate the updated design, we simulate 1000 rollout trajectories. The average return improves from -113.817 to -108.301, while the standard deviation decreases from 20.115 to 14.681 after the second optimization. Although the gain in expected return is modest, the significant reduction in variability demonstrates improved robustness of the closed-loop system.

The proposed framework follows a multi-generation design philosophy, where each generation involves co-optimization of system and control parameters informed by real-world data and model refinement. In this paper, we illustrate the initialization stage and the first three steps (first optimization, first physical implementation, and second optimization); however, the framework is inherently extensible. The number of generations required is problem-dependent and should be determined based on performance objectives and application constraints. Designers are encouraged to iteratively apply additional generations as needed to achieve desired levels of performance, reliability, and adaptability.

\section{Engineering Study: Active Suspension System}
\label{sec:EngineeringStudy}

Unlike passive suspensions, active suspension systems enhance passenger comfort and vehicle stability by dynamically adjusting damping forces. However, their design requires a synergetic collaboration between the physical and control domains to improve comfort and stabilize the vehicle under various disturbances. Although traditional Proportional–integral–derivative (PID) controllers are simple to implement, they suffer from significant limitations, including a lack of adaptability and the need for precise tuning of control gains. While MPC, which optimizes real-time control actions by predicting future system responses, improves performance by accounting for future disturbances (shown in Chapter 8 of \cite{tsai2023phd}), its applications are often limited by computational resources. 

Due to the need for rapid decision-making for active suspension systems, such systems are chosen as the case study for demonstrating the proposed framework with DRL controllers which can provide near-instantaneous adjustments to suspension dynamics by evaluating the explicit policy. Additionally, once deployed in a real-world setting, the system generates a large volume of operational data, which can be leveraged to continuously train and update the controller, improving performance over time. Moreover, suspension systems operate in highly uncertain and dynamic environments, where road conditions, driving speed, and external disturbances vary unpredictably. DRL is well-suited for handling such uncertainties by learning adaptive control strategies that generalize across diverse conditions.

\begin{figure}
    \centering
    \includegraphics[width=0.65\linewidth]{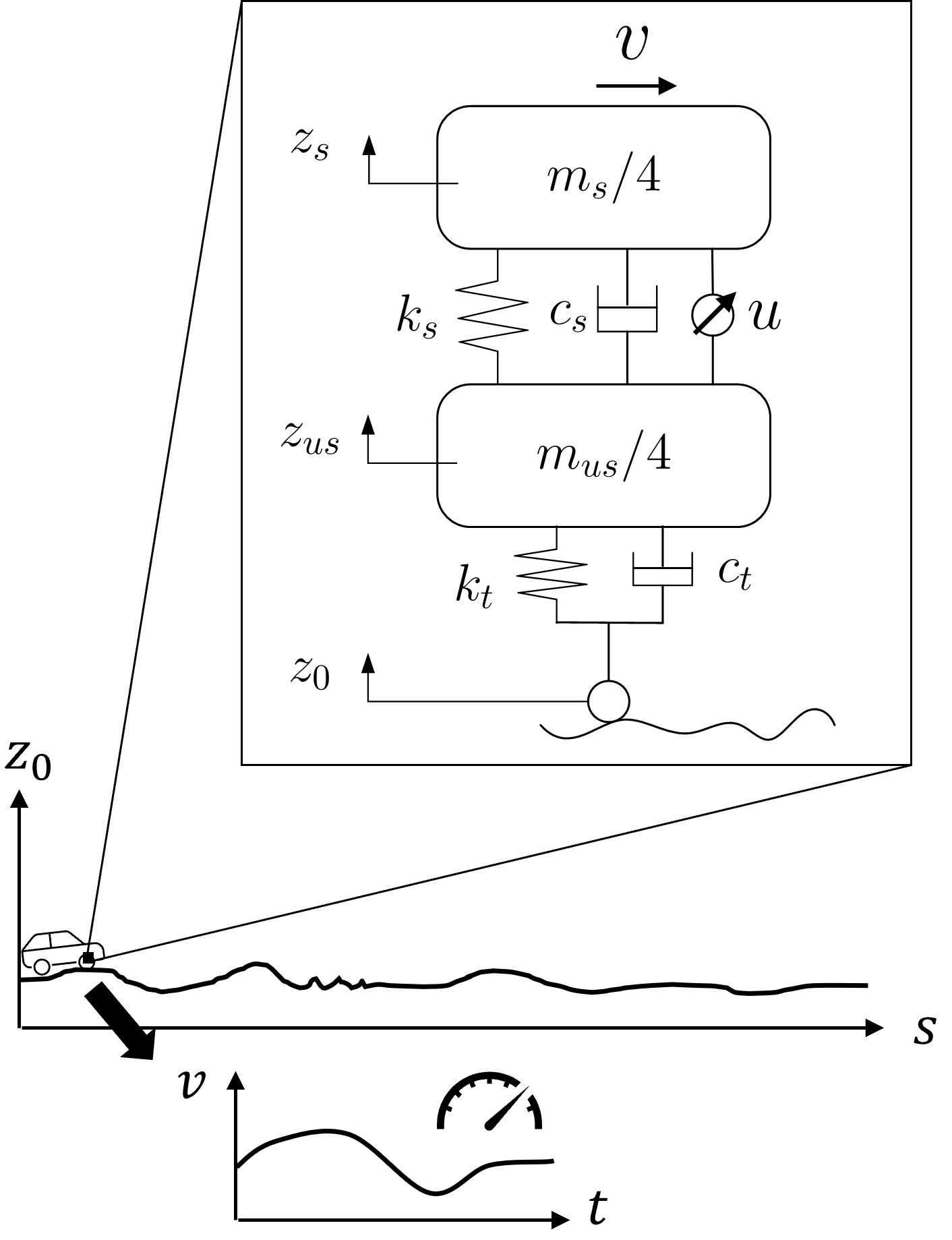}
    \caption{Quarter-car vehicle suspension model, modified from \cite{tsai2023phd}. Many uncertain factors affect the suspension system performance from the environment like road conditions and driving speed. Note that the coefficients for the spring and the damper, $k_s$ and $c_s$, are system parameters to be designed.}
    \label{fig:suspension_model_car}
\end{figure}

The quarter-car model of a vehicle suspension is shown in Fig.~\ref{fig:suspension_model_car}, where $m_s$ and $m_{us}$ are the sprung and unsprung masses, respectively, $z_s$ and $z_{us}$ are the vertical positions of the sprung and unsprung masses, respectively, $z_0$ is the elevation of the road which is the excitation source to the system, $k_t$ and $c_t$ are tire stiffness and damping constant, respectively, $k_s$ and $c_s$ are the coefficients of the spring and the damper (design variables for the physical system). The values of the parameters of the suspension system can be found in \cite{allison2014co}. The dynamic equation is defined by:
\begin{align}
        \underbrace{\begin{bmatrix}
            \dot{z}_{us}(t)-\dot{z}_0(t)\\
            \ddot{z}_{us}(t)\\
            \dot{z}_s(t)-\dot{z}_{us}(t)\\
            \ddot{z}_s(t)
        \end{bmatrix}}_{\dot{\mathbf{x}}(t)}&=
        \underbrace{\begin{bmatrix}
            0 & 1 & 0 & 0\\
            -\frac{k_{us}}{m_{us}} & -\frac{c_s}{m_{us}} & \frac{k_{us}}{m_{us}} & \frac{c_s}{m_{us}}\\
            0 & -1 & 0 & 1\\
            0 & \frac{c_s}{m_s} & -\frac{k_s}{m_s} & -\frac{c_s}{m_s}
        \end{bmatrix}}_{\mathbf{A_c}}
        \underbrace{\begin{bmatrix}
            {z}_{us}(t)-{z}_0(t)\\
            \dot{z}_{us}(t)\\
            {z}_s(t)-{z}_{us}(t)\\
            \dot{z}_s(t)
        \end{bmatrix}}_{\mathbf{x}(t)}\notag
        \\
        &+
        \underbrace{\begin{bmatrix}
        0\\
        -\frac{1}{m_{us}}\\
        0\\
        \frac{1}{m_{s}}
        \end{bmatrix}}_{\mathbf{B_c}}
        u(t)+\underbrace{\begin{bmatrix}
        -1\\
        \frac{c_t}{m_{us}}\\
        0\\
        0
        \end{bmatrix}}_{\mathbf{E_c}}
        \dot{z}_0(t).
    \label{eq:continuous_time_susp}
\end{align}

By selecting the sampling time $T=0.05$ sec, the continuous-time dynamic model in Eq.~(\ref{eq:continuous_time_susp}) can be converted in a discrete-time version:
\begin{equation}
    \mathbf{x}_{k+1}
    =
    \mathbf{A}
    \mathbf{x}_{k}
    +\mathbf{B}u_k+\mathbf{w}_k,
    \label{eq:discrete_time_susp}
\end{equation}
where $\mathbf{A}=e^{\mathbf{A_c}T}$ and $\mathbf{B}=\int_{0}^{T}e^{\mathbf{A_c}\tau}\mathbf{B_c}d\tau$. Although matrix $\mathbf{B_c}$ is independent on $k_s$ and $c_s$, matrix $\mathbf{B}$ can be affected by their values after discretization. The disturbance vector $\mathbf{w}_k$ is computed by:
\begin{equation}
    \mathbf{w}_k=\mathbf{E_d}\dot{z}_0(t_k),
\end{equation}
where $\mathbf{E_d}=\int_{0}^{T}e^{\mathbf{A_c}\tau}\mathbf{E_c}d\tau$ and $\dot{z}_0(t_k)$ is the first derivative of the elevation at time $t=t_k$, which can also be expressed as $\dot{z}_0(t_k)=\frac{dz_0(s)}{ds}v(t_k)$, where $s$ denotes distance. It means the disturbance $\dot{z}_0(t_k)$ depends on the road roughness and the vehicle speed $v$.

The task of the suspension system is to minimize vibrations and enhance ride comfort by regulating the state variables toward desired equilibrium points, effectively reducing oscillations caused by road roughness and other disturbances. Another important consideration is minimizing the control effort in order to reduce energy consumption. The reward function is defined by:
\begin{equation}
    r_{k+1}=-\mathbf{x}_{k+1}^\top\mathbf{Q}\mathbf{x}_{k+1}-10^{-6}u_k^2,
    \label{eq:reward_susp}
\end{equation}
where $\mathbf{Q}=\text{diag}([10,1,50,5])$.

\subsection{Step 1: Initial CCD Optimization}
In order to optimize the system with disturbances without prior knowledge of the environment, we assume the disturbance vector follows a Gaussian distribution, i.e., $\dot{z}_0(t_k)\sim\mathcal{N}(0,0.3^2)$. The initial values of the physical components are $k_s=27692.0$ N/m and $c_s=1906.5$ N$\cdot$s/m. For this case, we adopt similar DNN structures as described in the numerical example: fully connected feedforward networks with five hidden layers (16, 32, 32, 16, and 1 neurons) and \texttt{Tanh} activations. The input is 6-dimensional $(x_1,x_2,x_3,x_4, k_s, c_s)$, and the output is a scalar of control action $u$. The policy and value networks share the same architecture, with the policy consisting of two DNNs for the mean and standard deviation. The mean policy network is pre-initialized using 3,000 samples, while the standard deviation network is initialized with zero weights and bias values of 0.01.

We use 10000 epochs for the CCD optimization with the learning rate of $10^{-5}$. The learning rate was chosen to balance stability and convergence speed, preventing instability while enabling effective policy refinement, and 10,000 epochs were empirically selected to ensure sufficient training iterations for convergence without excessive computational cost, as fewer epochs led to underfitting while significantly more did not provide substantial additional improvements. The training process took approximately 3 hours on a system equipped with an AMD EPYC 7413 24-core CPU (2.65 GHz), 1 TB of RAM, running a 64-bit Linux operating system. The values have been updated to $k_s=31863.8$ N/m and $c_s=2067.1$ N$\cdot$s/m. The average returns before and after optimization are -87.429 and -82.106, respectively, while the standard deviations of these 1000 trajectories before and after optimization are 11.932 and 12.059. The improvement looks trivial at this stage because the system behavior is simple with a predefined form of disturbances and the initial design is close to optimum so far. In the next step as the system is placed in the real environment and the physical data is continuously collected, we can get more information about the environmental conditions and capture some underlying physics.

\subsection{Step 2: First Implementation (Generation 1)}
There are inherent discrepancies between the real system and its digital model, such as unknown environmental variations, and nonlinearities that our digital model does not fully capture. In the context of suspension systems, environmental disturbances come from unknown road conditions and driving behaviors. A road profile of several jumps, bumps, dents, and ramps with some noises over distances was created, while a speed profile with 15000 steps was generated as a function of time\footnote{The codes and data for the road and speed profiles are provided \href{https://github.com/TsaiYK/ControlCoDesign_DigitalTwin_RL.git}{https://github.com/TsaiYK/ControlCoDesign\_DigitalTwin\_RL.git}.\\}. The system nonlinearities exist in the mechanical components of the spring and damper. The dynamics of the real suspension system are defined as:
\begin{align}
    \dot{\mathbf{x}}(t)&=
    \mathbf{A_c}
    \mathbf{x}(t)
    +
    \mathbf{B_c}
    u(t)+\mathbf{E_c}
   \dot{z}_0(t)\notag\\
    &+
     \underbrace{\begin{bmatrix}
        0\\
        -\frac{1}{m_{us}}\left(k_{nl}x^3_3(t)+c_{nl}\left|x_4(t)-x_2(t)\right|\left(x_4(t)-x_2(t)\right)\right)\\
        0\\
        \frac{1}{m_s}\left(k_{nl}x^3_3(t)+c_{nl}\left|x_4(t)-x_2(t)\right|\left(x_4(t)-x_2(t)\right)\right)
    \end{bmatrix}}_{\text{Nonlinear terms of spring and damper}},
    \label{eq:continuous_time_susp}
\end{align}
where $k_{nl}=0.01k_s$ and $c_{nl}=0.01c_s$ represent the nonlinear coefficients of the spring and damper, whose values follow a ratio of the coefficients $k_s$ and $c_s$, and $\dot{z}_0(t)$ is provided given the profiles of road elevations and driving speeds. 

\begin{figure}
    \centering
    \includegraphics[width=0.95\linewidth]{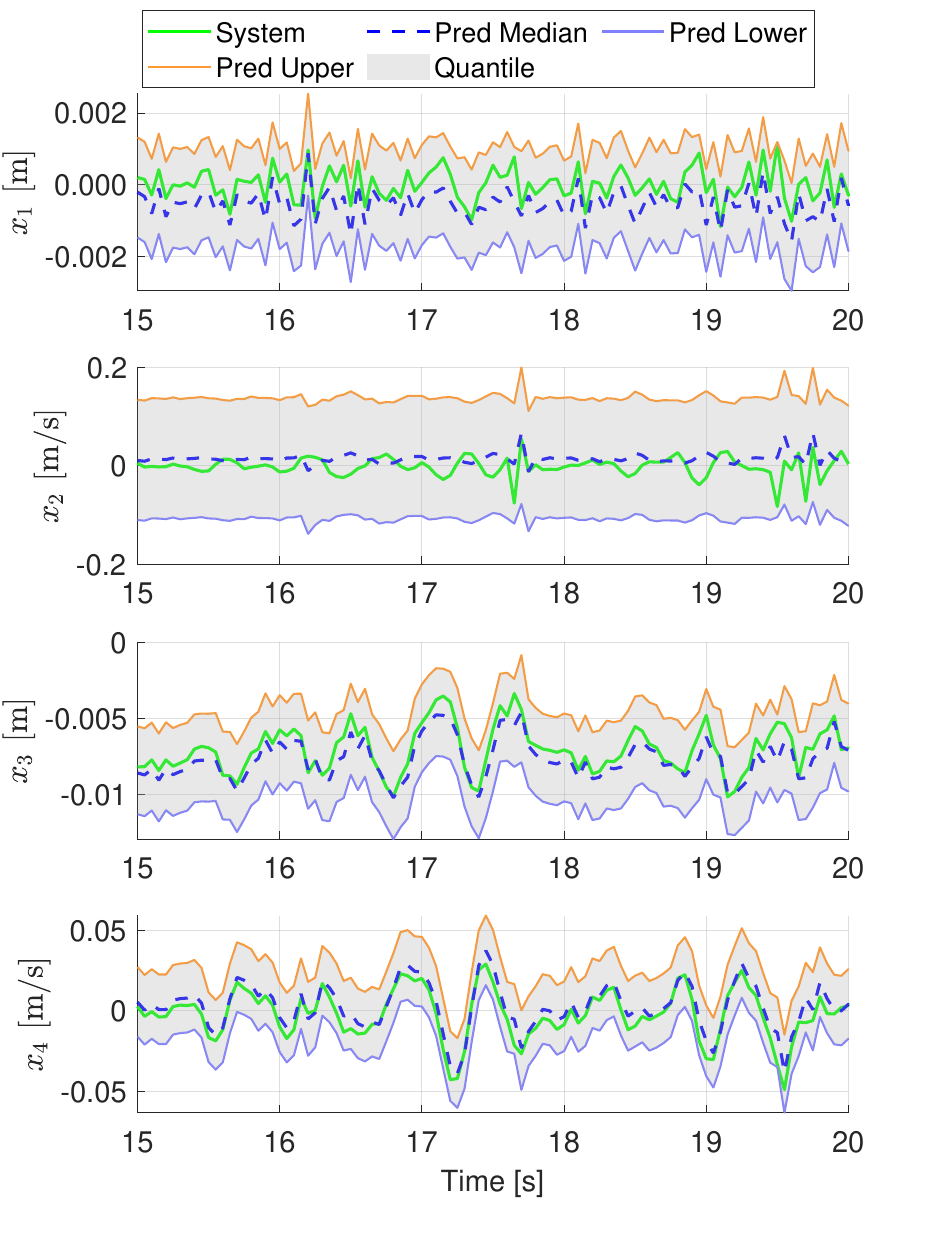}
    \caption{Visualization of the learned quantiles and the real system trajectories for the active suspension system.}
    \label{fig:quantile_susp}
\end{figure}

In practice, we can use sensors to measure the vertical movements and excitations from the road as well as the driving speeds in real-time. The collected data can be used for updating the RL policy and the digital model. A discrepancy model, defined in Eq.~(\ref{eq:quantile_numerical}), is trained using quantile regression. Figure~\ref{fig:quantile_susp} shows the learned quantiles with the real system trajectories, where the lower, median, and upper quantiles represent 10th, 50th, and 90th percentiles, respectively. The root mean square error for the validation dataset is 0.0343. The discrepancy model captures the uncertainty for most of the state variables with effectively learned quantiles; however, the quantile of the second state variable is larger than expected. This arises due to the sudden jumps of the vertical velocity of the tire coming from the rough road conditions. The presence of such abrupt disturbances results in an unbalanced dataset, where the frequency of extreme values is lower than that of regular conditions. Consequently, the model overestimates the actual uncertainty, leading to a more conservative prediction with a larger quantile.

\subsection{Step 3: Second CCD Optimization (Generation 2)}
The main difference between the first and second CCD optimizations is that the second CCD optimization uses the updated dynamic model with the learned quantile, shown in Eq.~(\ref{eq:updated_dynamic_model}), and the updated reward function, shown in Eq.~(\ref{eq:updated_reward_numerical}). With the physical data collected from Generation 1, the updated model more accurately captures environmental uncertainty through quantile learning, and there is no need for explicit assumptions about the disturbances.

\begin{figure}
    \centering
    \includegraphics[width=0.85\linewidth]{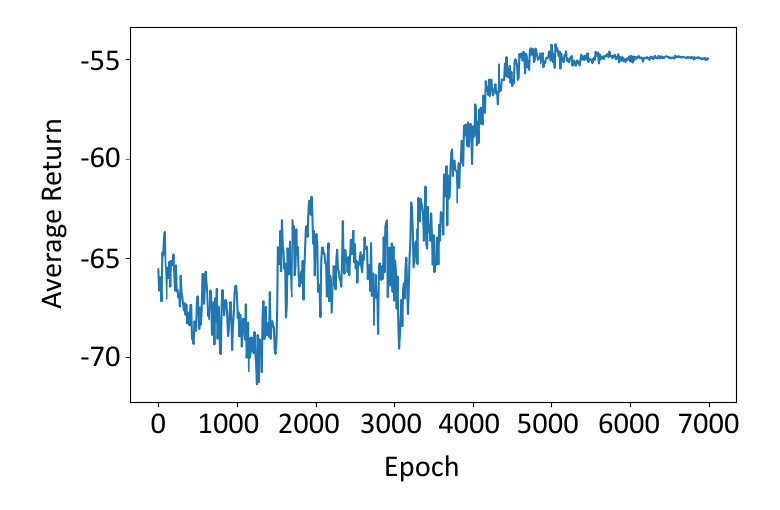}
    \caption{Training history for the active suspension system. The optimized values of the physical system parameters (the coefficients of the spring and damper) are $k_s = 32290.6$ N/m and $c_s=2072.6$ N$\cdot$s/m.}
    \label{fig:return_history_susp}
\end{figure}

The return history from the RL process exhibits an overall upward trend shown in Fig.~\ref{fig:return_history_susp}, indicating that the RL policy is progressively improving to maximize returns. The fluctuations observed in the trajectory are expected, as the decision-making process in RL is inherently stochastic. This variability arises from the trade-off between exploration and exploitation, which is crucial for operating in highly uncertain environments. For the active suspension system, the learning trend demonstrates a steady improvement in policy performance. {Furthermore, it is important to note that the results demonstrate a clear upward trend and indicate consistent performance improvement and convergence of the policy over time. The steady increase and eventual stabilization of the average return suggest that the proposed method effectively learns a near-optimal co-design solution. This behavior supports the reliability of the learning process and highlights the robustness of the approach in handling the simultaneous optimization of control and physical design variables under uncertainty.} The framework jointly optimizes the control strategy while adapting the physical parameters of the suspension system.

% \begin{table}[]
% \small
% \begin{center}
%     \caption{Mean and standard deviation (std) of the returns for Generation 1 design (Gen-1) and Generation 2 design (Gen-2).}
%     \begin{tabular}{ccc}
%     \Xhline{1pt}
%          & Mean of returns & Std of returns  \\\Xhline{1pt}
%          Gen-1 & -5.6018 & 0.2715\\\hline
%          {\bf Gen-2} & {\bf -2.2500} & {\bf 0.0686}\\\Xhline{1pt}
%     \end{tabular}
%     \label{tab:returns_step3_susp}
% \end{center}
% \end{table}

For the sake of simplicity, we refer to the solutions before and after the second CCD optimization as Generation 1 design (Gen-1) and Generation 2 design (Gen-2), respectively. To compare their performance, 200 samples of different initial conditions, we generate 200 samples of different initial conditions within the range $\mathbf{x}_{lb}=[-0.5, -2, -0.2, -1]^\top$ to $\mathbf{x}_{ub}=[0.5, 2, 0.2, 1]^\top$. Both designs are simulated for 100 steps (equivalent to 5 seconds), and their dynamic performance is evaluated by computing the returns for all 200 trajectories. Table \ref{tab:returns_step3_susp} presents the mean and standard deviation of the returns for Gen-1 and Gen-2 designs across the sampled initial conditions. Notably, the Gen-2 design exhibits significantly improved performance and robustness, as indicated by its lower mean and standard deviation of returns compared to Gen-1. This demonstrates that incorporating physical data and updating the digital model can substantially improve the design of autonomous systems.

\begin{table}[]
\begin{center}
    \caption{Mean and standard deviation (std) of the returns for Generation 1 design (Gen-1) and Generation 2 design (Gen-2).}
    \begin{tabular}{lll}
    \Xhline{1pt}
         & Gen-1 & {\bf Gen-2}  \\\Xhline{1pt}
         Mean of returns & -5.6018 & {\bf -2.2500}\\\hline
         Std of returns &  0.2715 & {\bf 0.0686}\\\Xhline{1pt}
    \end{tabular}
    \label{tab:returns_step3_susp}
\end{center}
\end{table}

% \begin{figure*}
%     \centering
%     \includegraphics[width=0.95\linewidth]{figures/susp_traj_trial2.pdf}
%     \caption{Third and forth states and control trajectories of (a) Gen-1 and (b) Gen-2 designs for the initial condition of $\mathbf{x}=[-0.2780, -1.0757, 0.0688, -0.9606]^\top$, where $\sigma_{ss}$ denotes the standard deviation of the steady-state response.}
%     \label{fig:susp_traj_trial1}
% \end{figure*}

\begin{figure*}
    \centering
	\begin{subfigure}{0.49\textwidth}
    	\includegraphics[width=\textwidth]{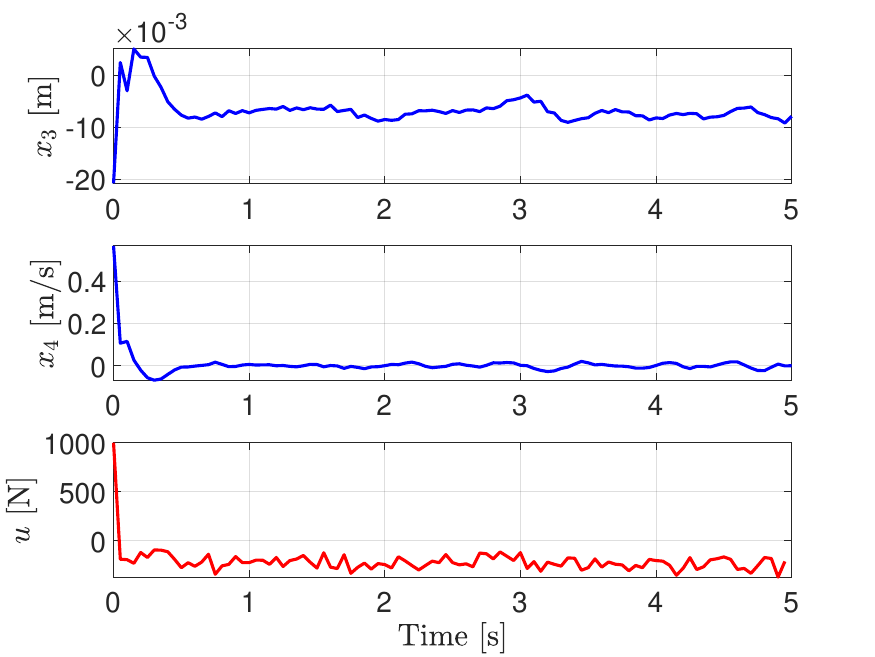}
    	\caption{Gen-1}
    	\label{fig:trial1_before}
	\end{subfigure}
    \hfill
	\begin{subfigure}{0.49\textwidth}
    	\includegraphics[width=\textwidth]{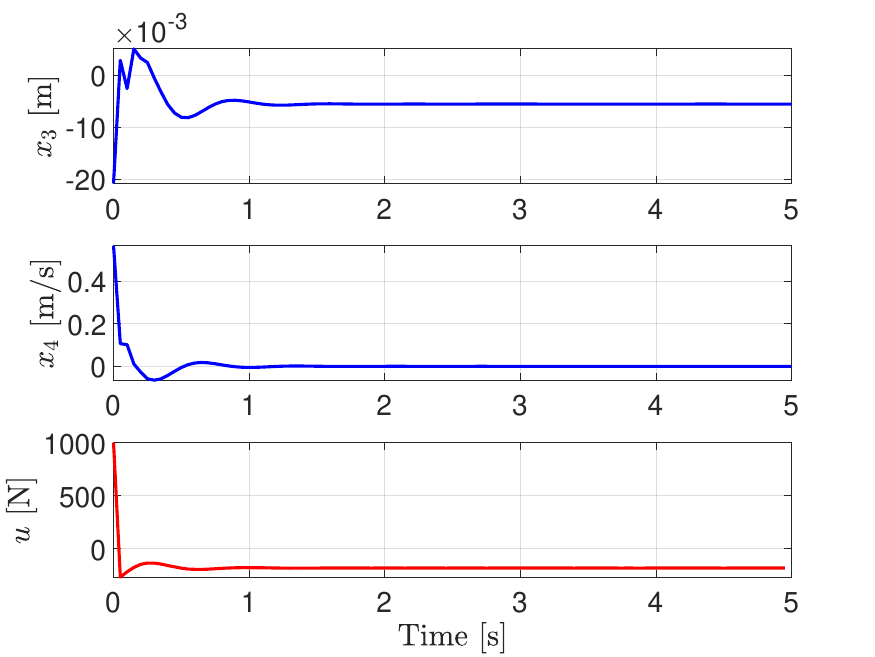}
    	\caption{Gen-2}
    	\label{fig:trial1_after}
	\end{subfigure}
    \begin{subfigure}{0.49\textwidth}
    	\includegraphics[width=\textwidth]{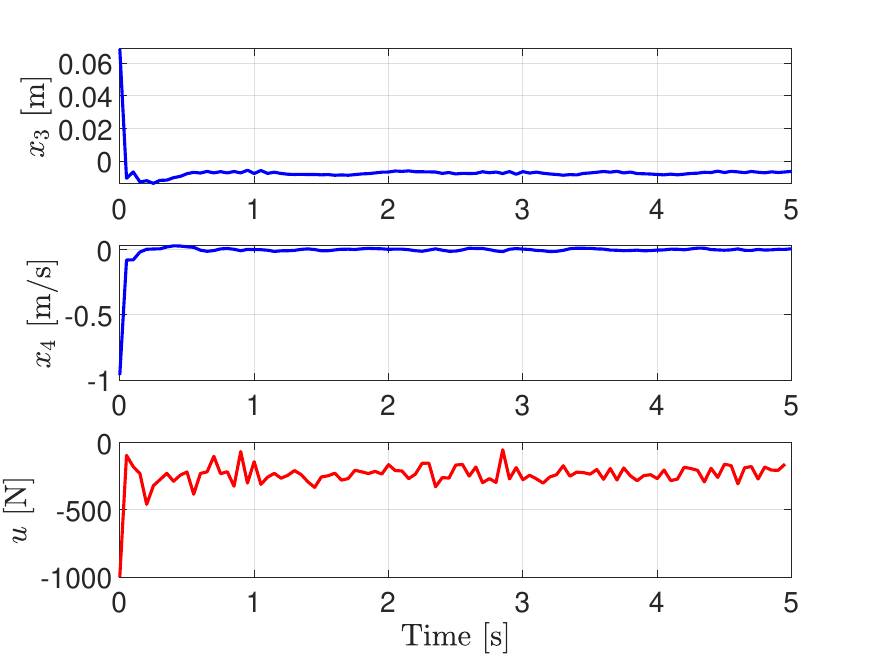}
    	\caption{Gen-1}
    	\label{fig:trial2_before}
	\end{subfigure}
    \hfill
	\begin{subfigure}{0.49\textwidth}
    	\includegraphics[width=\textwidth]{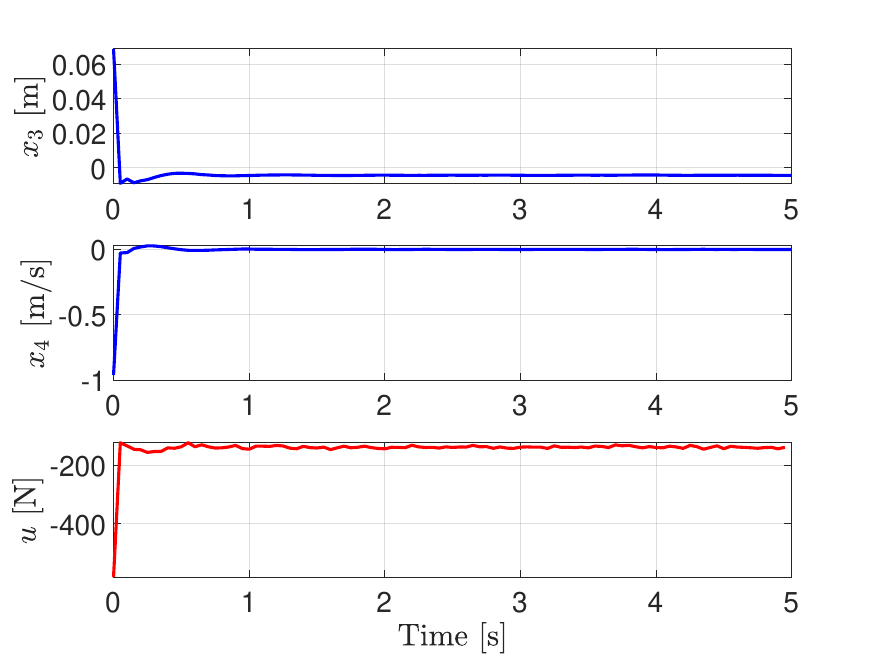}
    	\caption{Gen-2}
    	\label{fig:trial2_after}
	\end{subfigure}
    \begin{subfigure}{0.49\textwidth}
    	\includegraphics[width=\textwidth]{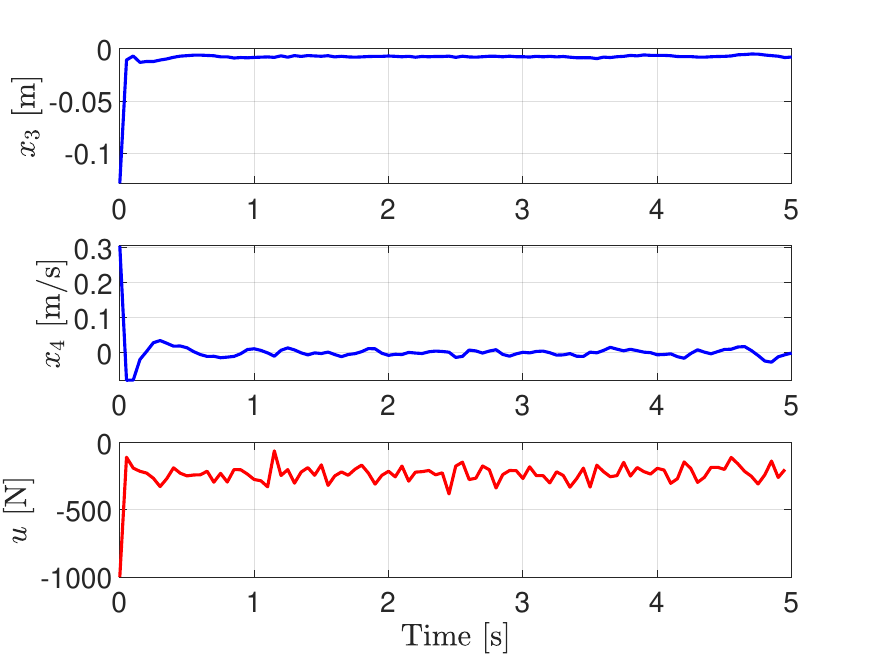}
    	\caption{Gen-1}
    	\label{fig:trial3_before}
	\end{subfigure}
    \hfill
	\begin{subfigure}{0.49\textwidth}
    	\includegraphics[width=\textwidth]{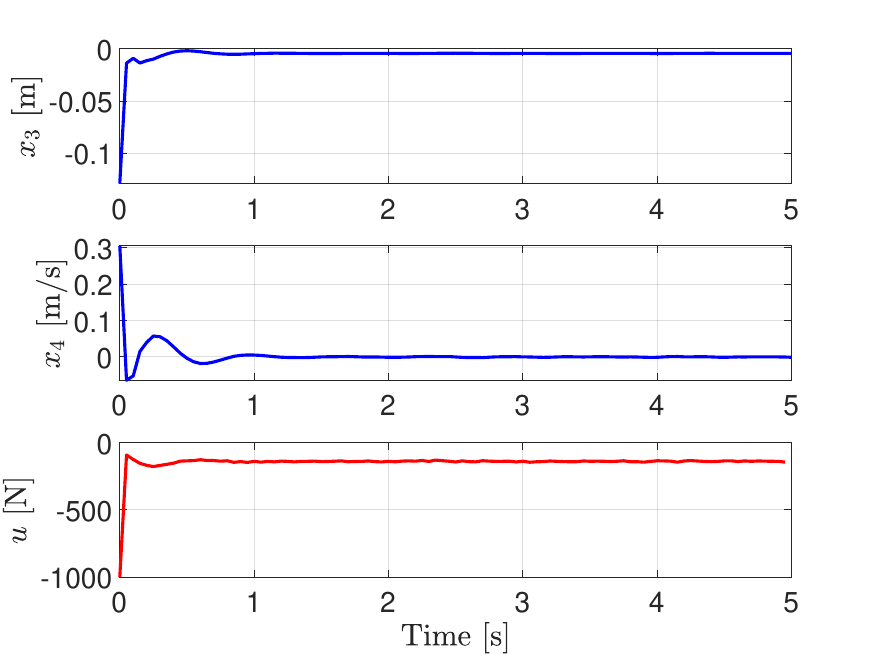}
    	\caption{Gen-2}
    	\label{fig:trial3_after}
	\end{subfigure}
    \caption{Third and fourth states and control trajectories of Gen-1 and Gen-2 designs for the initial conditions (a,b) $\mathbf{x}=[0.49,1.74,-0.02,0.57]^\top$, (c,d) $\mathbf{x}=[-0.28, -1.08, 0.07, -0.96]^\top$, and (e,f) $\mathbf{x}=[-0.40,1.20,-0.13,0.31]^\top$, where $\sigma_{ss}$ denotes the standard deviation of the steady-state response.}
    \label{fig:susp_traj_trial}
\end{figure*}

To qualitatively assess the improvements achieved through the second CCD optimization, their dynamic responses of the Gen-1 and Gen-2 designs (corresponding to the system before and after optimization, respectively) are compared in Fig.~\ref{fig:susp_traj_trial}. The responses are evaluated under different initial conditions. {The sharp increase in control input observed at the start of the trajectories ($u$ in Fig. \ref{fig:susp_traj_trial}) is due to the use of non-equilibrium initial conditions. This setup was intentionally chosen to evaluate and compare the dynamic response and stabilization behavior of different designs. In real-world scenarios, systems typically begin near equilibrium, and such abrupt control actions are less likely to occur.}

In Fig.~\ref{fig:susp_traj_trial}, the Gen-2 design exhibits significantly more stable responses with reduced control effort. Moreover, the control trajectories in Gen-2 are noticeably smoother, indicating improved efficiency. To further quantify the performance, the standard deviations of the steady-state responses, $\sigma_{ss}$, under three different initial conditions are computed and displayed in Table~\ref{tab:sigma_ss_susp}. Compared to Gen-1, Gen-2 achieves consistently lower values across both state variables and control actions. Notably, the reduction in the fourth state response, $x_4$, is especially important, as this variable directly relates to passenger comfort by mitigating oscillatory motion. At the same time, the control effort $u$ is substantially reduced, leading to smoother and more energy-efficient actuation. These results confirm that the Gen-2 design not only enhances comfort and robustness but also achieves significant energy savings in actuation.

% \begin{table}[]
% \small
% \begin{center}
%     \caption{Mean and standard deviation (std) of $\sigma_{ss}$ for Generation 1 design (Gen-1) and Generation 2 design (Gen-2).}
%     \begin{tabular}{cccccc}
%     \Xhline{1pt}
%          & & $x_1$ [m] & $x_2$ [m/s] & $x_3$ [m] & $x_4$ [m/s]  \\\Xhline{1pt}
%          \multirow{2}{*}{Gen-1} & Mean & 0.0007 & 0.0042 & 0.0017 & 0.0167  \\\cline{2-6}
%          & Std & 0.0002 & 0.0010 & 0.0004 & 0.0041\\\hline
%          \multirow{2}{*}{Gen-2} & Mean & 0.0005 & 0.0029 & 0.0012 & 0.0124\\\cline{2-6}
%          & Std & 0.0002 & 0.0013 & 0.0006 & 0.0055\\\Xhline{1pt}
%     \end{tabular}
%     \label{tab:returns_step3_susp}
% \end{center}
% \end{table}

% \begin{table}[]
% \small
% \begin{center}
%     \caption{Mean of steady-state responses, $\sigma_{ss}$, for Generation 1 design (Gen-1) and Generation 2 design (Gen-2).}
%     \begin{tabular}{cccccc}
%     \Xhline{1pt}
%          & $x_1$ [m] & $x_2$ [m/s] & $x_3$ [m] & $x_4$ [m/s] & $u$ [N]  \\\Xhline{1pt}
%          {Gen-1} & 0.0007 & 0.0042 & 0.0017 & 0.0166 & 56.57\\\hline
%          {\bf Gen-2} & {\bf 0.0005} & {\bf 0.0029} & {\bf 0.0012} & {\bf 0.0124} & {\bf 8.26}\\\Xhline{1pt}
%     \end{tabular}
%     \label{tab:sigma_ss_susp}
% \end{center}
% \end{table}

\begin{table}
\centering
\caption{Standard deviations of the steady-state trajectories $\sigma_{ss}$ for Gen-1 and Gen-2 under three different initial conditions.}
\begin{tabular}{llll}
\Xhline{1pt}
\textbf{Condition} & \textbf{Metric} & \textbf{Gen-1} & \textbf{Gen-2} \\\Xhline{1pt}
\multirow{3}{*}{1} & $\sigma_{ss}$ of $x_3$ (m)   & 0.0025  & 0.0020  \\
& $\sigma_{ss}$ of $x_4$ (m/s) & 0.0222  & {\bf 0.0194}  \\
& $\sigma_{ss}$ of $u$ (N)     & 59.4364 & {\bf 13.1011} \\
\hline
\multirow{3}{*}{2} & $\sigma_{ss}$ of $x_3$ (m)   & 0.0014  & 0.0008  \\
& $\sigma_{ss}$ of $x_4$ (m/s) & 0.0142  & {\bf 0.0064}  \\
& $\sigma_{ss}$ of $u$ (N)     & 60.8681 & {\bf 4.9897}  \\
\hline
\multirow{3}{*}{3} & $\sigma_{ss}$ of $x_3$ (m)   & 0.0013  & 0.0017  \\
& $\sigma_{ss}$ of $x_4$ (m/s) & 0.0151  & {\bf 0.0139}  \\
& $\sigma_{ss}$ of $u$ (N)     & 54.4229 & {\bf 8.9099}  \\
% \hline
% \multirow{3}{*}{4} & $\sigma_{ss}$ of $x_3$ (m)   & 0.0015  & 0.0018  \\
% & $\sigma_{ss}$ of $x_4$ (m/s) & 0.0156  & 0.0168  \\
% & $\sigma_{ss}$ of $u$ (N)     & 59.2047 & 11.1066 \\
\Xhline{1pt}
\end{tabular}
\label{tab:sigma_ss_susp}
\end{table}

% \begin{figure*}
%     \centering
% 	\begin{subfigure}{0.49\textwidth}
%     	\includegraphics[width=\textwidth]{figures/susp_traj_trial1.pdf}
%     	\caption{}
%     	\label{fig:susp_traj_trial1}
% 	\end{subfigure}
%     \hfill
% 	\begin{subfigure}{0.49\textwidth}
%     	\includegraphics[width=\textwidth]{figures/susp_traj_trial1.pdf}
%     	\caption{}
%     	\label{fig:susp_traj_trial1}
% 	\end{subfigure}
%     \caption{State and control trajectories for active suspension systems before and.}
%     \label{fig:susp_traj}
% \end{figure*}

\section{Conclusion and Future Work}
\label{sec:Conclusion}
This work presents a lifecycle-oriented Control Co-Design (CCD) framework that integrates Digital Twins (DTs) and Deep Reinforcement Learning (DRL) to address the challenge of designing dynamic systems under unpredictable uncertainties. The proposed approach extends CCD beyond single-cycle optimization by enabling DTs to evolve through real-time sensing, model updating, and adaptive re-optimization across successive generations. DRL accelerates real-time decision-making and controller adaptation, while multi-generation design leverages operational data from each deployed system to refine DT models, improve uncertainty quantification through quantile regression, and inform the redesign of both physical and control components for subsequent generations. Together, these elements ensure that system performance and robustness improve progressively over the lifecycle. The key contributions of this work are: (1) extending CCD into a multi-generation framework that explicitly incorporates the full product lifecycle, (2) leveraging DTs to enable continuous model evolution and data-driven decision-making, and (3) employing DRL to accelerate adaptive control and enhance real-time responsiveness. 

The proposed framework is demonstrated on an active suspension system, where DRL and automatic differentiation are employed to co-optimize the system parameters for the mechanical components and control policies. Quantile regression is used to learn the environmental uncertainty from physical data on road conditions and driving speeds after the deployment. The results indicate that the optimized solution after the second CCD optimization, which incorporates the updated model and control policy, achieves better dynamic responses, smoother control trajectories, and improved robustness compared to the initial design. A key advantage of this framework is its ability to co-optimize the physical system and its controller to ensure that performance and robustness improve over successive generations. 

{One limitation of the proposed approach is the need for many training epochs to ensure stable convergence when jointly optimizing the system and policy. This is due to the complexity of co-adaptation and the goal of achieving generalizability. Future work will explore integrating model predictive control (MPC) to guide long-horizon planning and improve sample efficiency.} Another future work will investigate more case studies of active suspension systems across diverse scenarios, aiming to optimize designs for different vehicle types, driving styles, and regional conditions. Given that optimal system configurations vary based on operational requirements, our proposed framework’s ability to autonomously learn from the environment and co-optimize both physical components and control policies is particularly valuable. One key direction worth investigating is how we achieve optimal designs for varying driving speed profiles and road conditions. Furthermore, while this work focuses on active suspension, the methodology is generalizable to other engineering applications, such as wind turbines and autonomous vehicles, where dynamic interactions between control and physical system design play a crucial role.

\section*{Acknowledgement}
We are grateful for the grant support from the National Science Foundation (NSF)’s Engineering Research Center for Hybrid Autonomous Manufacturing: Moving from Evolution to Revolution (ERC-HAMMER) and NSF's FMRG: Manufacturing ADvanced Electronics through Printing Using Bio-based and Locally Identifiable Compounds (MADE-PUBLIC), under Award Numbers EEC-2133630 and CMMI-2037026, respectively. Yi-Ping Chen also acknowledges the Taiwan-Northwestern Doctoral Scholarship to support his doctoral study. We also thank Dr. Anton van Beek at University College Dublin for his feedback and insightful discussions.

\selectlanguage{english} 

% When you drop the [french] option, delete your .aux file as well, since [french] sets ":" as an active character.

%%%%%%%%%%%%%  BIBLIOGRAPHY  %%%%%%%%%%%%%%%%%%%%%%%%%%%%%%%%%%%%%%%%%

%\nocite{*} %% <=== delete this line - unless you wish to typeset the entire contents of your .bib file.

\bibliographystyle{main}   %% .bst file that follows ASME journal format. Do not change.

\bibliography{main} %% <=== change this to name of your bib file

%%%%%%%%%%%%%%%%%%%%%%%%%%%%%%%%%%%%%%%%%%%%%%%%%%%%%%%%%%%%%%%%%%%%%%

%% To omit final list of figures and tables, use the class option [nolists]

\end{document}